\documentclass[11pt, a4paper, logo, internal]{dm}
\usepackage{graphicx}%
\usepackage{multirow}%
\usepackage{amsmath,amssymb,amsfonts}%
\usepackage{amsthm}%
\usepackage{mathrsfs}%
\usepackage[title]{appendix}%
\usepackage{xcolor}%
\usepackage{textcomp}%
\usepackage{manyfoot}%
\usepackage{booktabs}%
\usepackage{algorithm}%
\usepackage{algorithmicx}%
\usepackage{algpseudocode}%
\usepackage{listings}%
\usepackage{hyperref}
\usepackage{lineno}

\theoremstyle{thmstyleone}%
\theoremstyle{thmstyletwo}%

\theoremstyle{thmstylethree}%
\usepackage{mathtools}
\usepackage[dvipsnames]{xcolor}
\usepackage[numbers, sort&compress, round]{natbib}
\usepackage{booktabs}
\usepackage{graphicx}
\usepackage{xfrac}
\usepackage{bbm}
\usepackage{changes}
\usepackage{rotating}
\usepackage{wrapfig}

\usepackage[most]{tcolorbox}
\usepackage{xparse}
\usepackage{adjustbox}
\usepackage{xspace}
\usepackage{changepage}
\usepackage{enumitem}
\usepackage{pifont}
\usepackage{ulem}
\usepackage{tocloft}
\usepackage[toc]{multitoc}
\usepackage{etoc}
\usepackage{dsfont}
\usepackage{dm-colors}
\usepackage{multirow}
\usepackage{minted}
\usepackage{caption}
\usepackage{soul}
\usepackage{float}
\usepackage{svg}
\usepackage{adjustbox}
\usepackage{setspace}
\svgsetup{inkscapelatex=false}
\usepackage{silence}  
\newcommand{\hide}[1]{}

\newcommand{\method}{SpatialAgent\,}

\pdftrailerid{redacted}

\def\method{MixTIME}

\title{Predicting Immune Biomarkers with MultiModal Mixture-of-Expert Pathology Foundation Models Empowers Precision Oncology}

\begin{document}

\author[1,2,14,*]{Tianyu Liu}
\author[3,14]{Ziqing Wang}
\author[4,14]{Zhaokang Liang}
\author[5,6,7,8]{Tong Ding}
\author[9,15]{Peter Humphrey}
\author[9,15]{Lorraine Colón-Cartagena}
\author[10,11,15]{Emily Ling-Lin Pai}
\author[12,15]{Kenneth Tou En Chang}
\author[9]{Mohamed Kahila}
\author[13]{Jonathan Chong Kai Liew}
\author[1]{Tinglin Huang}
\author[1]{Rex Ying}
\author[3]{Kaize Ding}
\author[5,6,7]{Faisal Mahmood}
\author[2,4,*]{Wengong Jin}

\affil[1]{Program of Computational Biology and Bioinforamtics, Yale University, USA}

\affil[2]{Broad Institute of MIT and Harvard, USA}

\affil[3]{Department of Statistics and Data Science, Northwestern University, USA}

\affil[4]{Department of Computer Science, Northeastern University, USA}

\affil[5]{Department of Pathology, Mass General Brigham, Harvard Medical School, USA}

\affil[6]{Cancer Program, Broad Institute of Harvard and MIT, USA}

\affil[7]{ Data Science Program, Dana-Farber Cancer Institute, USA}

\affil[8]{ Harvard John A. Paulson School of Engineering and Applied Sciences, Harvard University, USA}

\affil[9]{Department of Pathology, Yale University, USA}

\affil[10]{Department of Anatomic Pathology and Laboratory Medicine, Hospital of the University of Pennsylvania, USA}

\affil[11]{Department of Pathology and Laboratory Medicine, University of California, San Francisco, USA}

\affil[12]{Department of Pathology and Laboratory Medicine, KK Women's and Children's Hospital, SGD}

\affil[13]{Department of Biostatistics, Epidemiology and Informatics, Perelman School of Medicine, University of Pennsylvania, USA}

\affil[14]{These authors contributed equally to this work.}

\affil[15]{These authors contribute equally to this project as human experts}

\affil[*]{Corresponding Authors.}




\begin{abstract}
Predicting immune biomarkers associated with the tumor immune microenvironment (TIME) is critical for advancing precision oncology, yet existing approaches are largely limited to single-image modalities and suffer from insufficient resolution and incomplete utilization of complementary clinical and biological information. Here we introduce MixTIME, a multimodal foundation model that leverages a mixture-of-experts (MoE) architecture to integrate pathology foundation models trained across distinct modalities: image-only (UNIv2), image-text (CONCHv1.5), and image-transcriptomic (STPath) representations for pixel-level and slide-level prediction of multiplex immunofluorescence (mIF) protein expression from hematoxylin and eosin (H\&E) whole-slide images. MixTIME employs a learnable router to dynamically weight expert contributions and is trained with a distribution- and tendency-aware loss function. Benchmarked on two datasets of different scales, MixTIME achieves state-of-the-art performance across 17 protein markers as measured by correlation metrics. The predicted mIF profiles substantially enhance downstream tasks, including spatial domain identification, survival prediction, and AI-assisted pathology report generation validated by expert pathologists from multiple institutes across the world. Furthermore, MixTIME enables longitudinal tracking of protein expression dynamics across clinical time points and reveals protein–gene interaction patterns linked to drug resistance and immune suppression in tumor microenvironments. Collectively, MixTIME provides a scalable framework for multimodal biomarker discovery and clinical translation in computational pathology.
\end{abstract}




\maketitle
\section{Introduction}
In oncology research, predicting the biomarkers associated with tumor immune microenvironment (TIME) allows us to better understand cancer progression, tumor growth, invasion, metastasis, and response to cancer treatment \cite{nagarsheth2017chemokines,wang2024immunotherapy,bagaev2021conserved}. Among the candidates of these markers, along with the backbone hematoxylin and eosin (H\&E) images \cite{chen2017computer}, proteomics has always been a vital source of information. Previous studies have demonstrated that observing and analyzing individual or multiple proteins can help us better characterize the TIME \cite{binnewies2018understanding, valanarasu2026multimodal}, while other modalities such as transcriptomic profiles \cite{huang2025stpath,liu2026leveraging,chen2025visual} as well as textual annotation \cite{lu2024visual} can also provide multi-angle information to help us better model TIME. More specifically, to simultaneously measure the expression levels and localization of multiple proteins, multiplex immunofluorescence (mIF) has been introduced as a powerful protein profiling method on the targeted tissue, while preserving the spatial architecture. Therefore, combining mIF measurement with other modalities represents a highly promising avenue for advancing both molecular and digital pathology research \cite{odell2013immunofluorescence}.

A major challenge in the past was the limitation of data volume. Due to the prohibitively high computational cost of mIF predictions, it was difficult to scale up datasets sufficiently to train reliable models. However, the recent expansion and release of public datasets have significantly altered this situation. Researchers can now leverage public mIF measurement paired with H\&E images (slides) and train reliable models to make predictions based on new slides. However, current methods also have some limitations. For example, recent approaches \cite{valanarasu2026multimodal, balezo2026miphei, wu2025rosie, li2026ai} only focus on predicting mIF with a single image modality. Moreover, different methods also have different resolutions, while low predictive accuracy may compromise the effectiveness of mIF in modeling the tumor microenvironment. We also note that researchers ignore many complementary modalities, such as gene expression profiles, which could be measured by spatial transcriptomic (ST) technology, and patch-level annotation produced by human experts or AI agents, to improve model performances and extend the applications to more scenarios. Therefore, we recognize the necessity of proposing a new algorithm to tackle these emergent problems.

Here we introduce \method{}, a multimodal foundation model designed for predicting mIF with a high resolution (pixel-level continuous outcome). \method{} leverages the mixture-of-expert (MOE) architecture \cite{shazeer2017outrageously} to integrate information from pathology foundation models (PFMs) trained with different modalities, including image-only (e.g. UNIv2 \cite{chen2024towards}), image-text (e.g. CONCHv1.5 \cite{lu2024visual}), and image-gene (e.g. STPath \cite{huang2025stpath}). \method{} is trained with a distribution-aware and tendency-aware loss function and can better model the real mIF strength across multiple channels. \method{} not only can predict mIF accurately and provide external information for assigned H\&E images across different scales (both patch and whole-slide image), but also explains the contributions of different experts in the prediction process and reveals the biological connections across multiple modalities. We demonstrate that \method{} achieves the state-of-the-art performances across multiple downstream applications, such as protein strength prediction, spatial domain identification and survival prediction. \method{} also imputes unmeasured information for patient samples across various time points and generates high-quality medical reports validated by pathologists. Overall, \method{} contributes to important applications in both biomolecular and clinical aspects. 

\section{Results}
\begin{figure}
    \centering
    \includegraphics[width=1\linewidth]{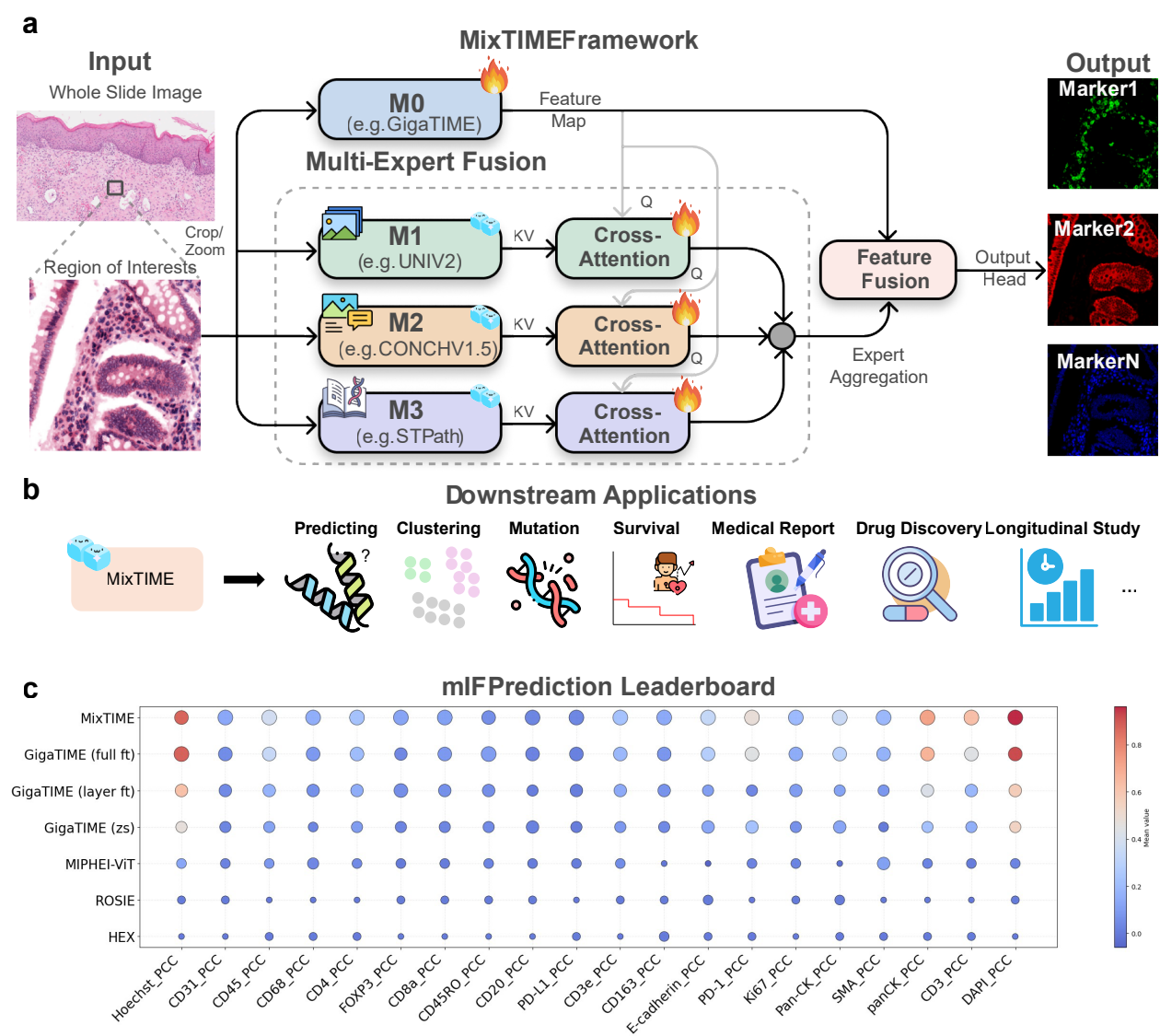}
    \caption{The landscape of \method{} for mIF prediction and multi-modal-enhanced downstream applications. (a) Model architecture. We leverage the mixture-of-expert model for prediction. (b) Downstream applications. (c) A bubble plot-based leaderboard for mIF prediction across two different datasets and the corresponding protein markers. A warmer color represents a higher score, and a large bubble represents a lower-ranked method.}
    \label{fig:overview}
\end{figure}

\textbf{Method overview.} Different from the traditional Image-to-X model, we build \method{} based on a mixture-of-expert (MOE) architecture to leverage the contribution of pathology foundation models (PFMs) trained with different modalities. We utilize image-based PFMs, image-text-based PFMs, image-ST-based PFMs, and a pre-trained mIF predictor as expert models, and design a router-enhanced method to integrate information flow from different base models to predict mIF information based on new HE images. \method{} can predict mIF in both patch-level and whole-slide-level resolutions, and the predicted mIF can be treated as an enhanced modality for several downstream applications, including spatial domain identification (clustering), survival prediction, medical report generation, etc. We also propose new tasks by using mIF as a biomarker for drug resistance detection and disease-state diagnosis. The overall model design with illustration of downstream applications, as well as the overall performance of benchmarking mIF prediction, can be found in Figures \ref{fig:overview} (a) and (b).

\textbf{Comprehensive benchmark demonstrates the contribution of integrating different modalities with the mixture-of-expert architecture.} A major advantage of \method{} is that it leverages prior information from multiple modalities, thereby enabling us to infer the intensity of mIF. This is an advantage that previous work lacked. Our architecture does not simply forcefully integrate or sum information from different modalities; instead, it uses the MoE architecture to quantify the contribution of each modality to the prediction. This allows the model to learn, during training, how to allocate weights for utilizing different modalities, thereby achieving optimal results. We also improve the design of the loss function to take care of the balance between magnitude-level prediction and tendency-level prediction. Our experiments on datasets (HEMIT \cite{bian2024hemit}, a smaller dataset, and ORION \cite{lin2023high}, a large dataset) of two different scales demonstrate that our current design has advantages. Our evaluation metrics follow the settings in GigaTIME \cite{valanarasu2026multimodal}, including pixel-level Pearson Correlation Coefficient (PCC) and Spearman Correlation Coefficient (SCC). Our baseline methods include GigaTIME (zs, where ``zs" means no fine-tuning) and its two variations (GigaTIME (layer ft, where ``layer ft" means we only tune the parameters in the last layer) and GigaTIME (full ft, where ``full ft" means we tune the parameters of all layers)), and the finetuned versions of HEX \cite{li2026ai}, MIHPEI-VIT \cite{balezo2026miphei}, and ROSIE \cite{wu2025rosie}. Details of baseline information can be found in the Methods section. 

\begin{figure}
    \centering
    \includegraphics[width=1\linewidth]{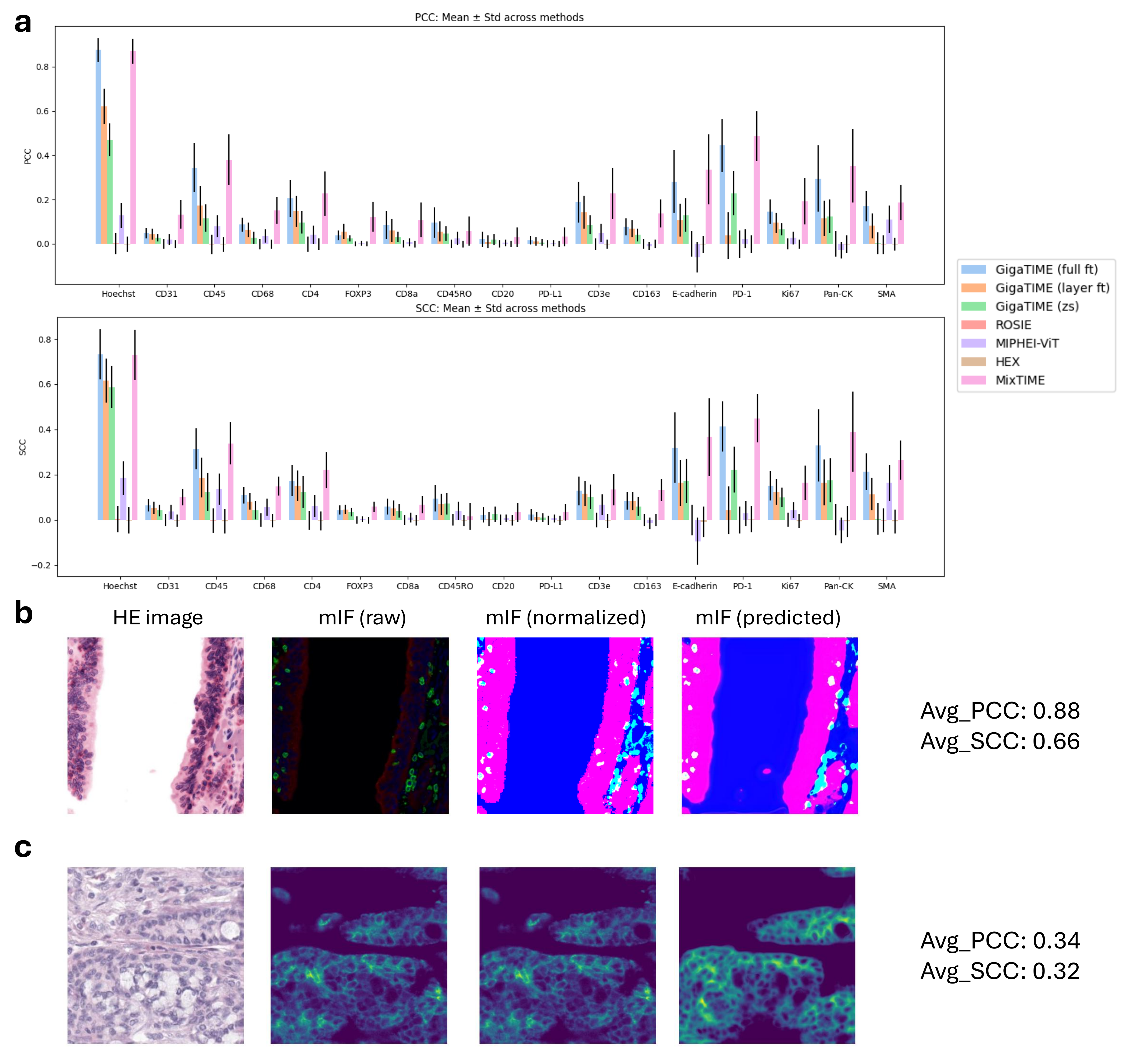}
    \caption{Results of mIF prediction comparison. (a) Mean value and standard deviations of PCCs and SCCs based on ORION dataset across 17 protein markers. (b) A case study based on HEMIT dataset with annotated image pixel information, raw mIF strength, and predicted mIF of three proteins. (c) A case study based on ORION dataset with annotated image pixel information, raw mIF strength, and predicted mIF of PD-1.}
    \label{fig:mifresult}
\end{figure}

According to Figure \ref{fig:overview} (b), if we take into account the predictive performance of different biomarkers across the two datasets, we can see that \method{} ranks highest in the comparison based on PCCs. We also zoom in and study the comparison of different datasets with PCCs and SCCs, shown in Supplementary Figure \ref{supfig:hemit result} for HEMIT and Figure \ref{fig:mifresult} (a) for ORION. For the HEMIT dataset, the advantages of \method{} across three biomarkers are clear, and since it is designed to predict the expression at the pixel level, it also does not need to run an approximation method to extend the predictions from the patch level to the pixel level (such as HEX and ROSIE), and thus has a higher resolution. For the ORION dataset, \method{} performs better than the selected baseline methods in most of the biomarkers, except the background protein Hoechst. Moreover, for proteins that are difficult to predict, our method does not significantly improve prediction accuracy. We believe this is primarily related to the quality of the data in the mIF measurements. In particular, if the protein expression level is already relatively low, the prediction will be even more difficult. For the patch with relatively better prediction performance across two datasets, we visualize them in Figures \ref{fig:mifresult} (b) and (c) as case studies. Here we find that our predicted protein expression levels are still quite close to the measured expression levels. \method{} also supports whole-slide-level mIF prediction, shown in Supplementary Figure \ref{supfig:wholeslide}. Overall, our evaluation generally indicates that using MoE to modulate the contribution of information from different modalities to training, combined with a better loss function, can improve prediction performance.

To understand why this approach improves the performance of mIF predictions, we also conducted rigorous ablation experiments using the HEMIT dataset. Supplementary Figure \ref{supfig:abla_study} (a) shows our comparison of MoE selection based on the L1 loss \cite{pytorchframework}. Here we find that using all of the experts has the best performance evaluated based on both PCCs and SCCs. Moreover, the input from STPath's generations should be gene expression profiles rather than low-dimensional embeddings. If we reduce the number of experts—that is, the sources of the input modalities—the prediction performance will decline, which is in line with our expectations. Supplementary Figure \ref{supfig:abla_study} (b) shows our ablation studies for adjusting loss functions. Here we find that using the current mixed loss functions based on PCC loss and MSE loss gives us the best performance, which surpasses other choices that either capture magnitude-level similarity or correlation-level similarity. The theoretical and experimental basis for this loss function has been discussed in previous work on multimodal biomarker expression prediction tasks \cite{liu2025unicorn}.

\begin{figure}
    \centering
    \includegraphics[width=1\linewidth]{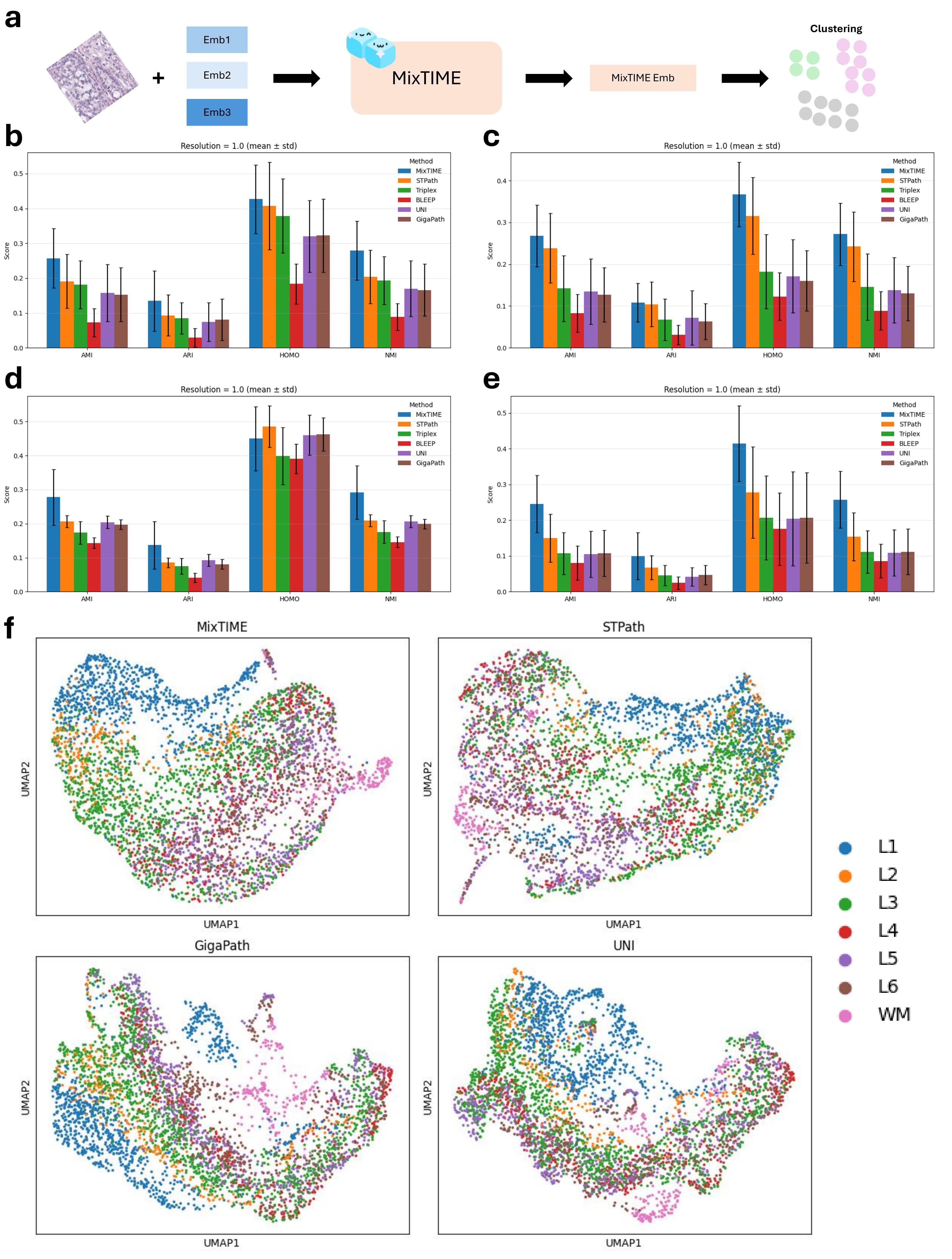}
    \caption{Benchmarking analysis based on the clustering task. (a) Workflow of \method{} in this application. (b)-(e) Clustering metrics across four different datasets with 1.0 resolution. (f) Case study with UMAPs from different methods based on the selected dataset.}
    \label{fig:clustering}
\end{figure}

\textbf{Extra mIF information enhances clustering and survival prediction.} With the mIF data now available, we have filled another gap in our pathological information, enabling us to use this predicted modality to support analysis for downstream tasks, particularly those related to biological features and clinical needs. Here, we explore the spatial domain identification task discussed in STPath. This task is derived from a spatial transcriptomics annotation project, in which researchers need to identify spot types based on images, spatial locations, and gene expression data. \method{} can predict the mIF information first and then produce embeddings enriched with transcriptomics, image, text, and mIF measurements for clustering. This pipeline is illustrated in Figure \ref{fig:clustering} (a), and our datasets used for benchmarking analysis come from STImage1k4M \cite{chen2024stimage}, which has human-annotated spatial domains with multiple samples. Since this is an unsupervised clustering task, we also use clustering-related metrics to evaluate the model's performance, including Adjusted Rand Index (ARI), Adjusted Mutual Information (AMI), Homogeneity (HOMO), and Normalized Mutual Information (NMI) \cite{pedregosa2011scikit}. Details of these metrics are included in the Methods section. Our baseline methods include STPath, Triplex \cite{chung2024accurate}, BLEEP \cite{xie2023spatially} (these three methods integrate both transcriptomic information and image information for clustering), UNIv2 \cite{chen2024towards}, and GigaPath \cite{xu2024whole} (these methods only use image information for clustering). Since \method{} has been shown as the optimal solution in mIF prediction, we do not consider other mIF predictors in the baseline settings. 

Figures \ref{fig:clustering} (b)-(e) show the clustering performances measured with different metrics across four selected datasets. Our results show that \method{} performs better than other baseline methods across all metrics within these four datasets, and thus \method{}'s predicted biomarkers can benefit spatial domain identification for samples from different tissues and disease states. The variance of \method{} is also comparable with other methods, and thus we do not sacrifice the robustness for the improvement of domain identification quality. We also visualize a set of samples from the human brain subset in STImage1k4M based on UMAPs colored by human-annotated spatial domains. Figure \ref{fig:clustering} (f) shows that \method{}'s embeddings perform better in identifying spots from L6 (red) and WM (purple), where other methods, such as UNIv2, tend to cluster different spots together.

We also tune the resolution used in the Leiden algorithm to make a fair comparison. In the main text, we report the maximal clustering performances of different baseline methods (resolution$=$1.0), and other resolutions are reported in Supplementary Figures \ref{supfig:gse},\ref{supfig:erickson},\ref{supfig:brain},\ref{supfig:andersion}. Even in the conditions of different resolutions, we can still identify the contributions of \method{} supported by consistent improvement.

Inspired by STPath \cite{huang2025stpath}, we also test \method{} with the weakly supervised survival prediction task. Four datasets, including MBC, SURGEN, HNSC, and LUAD from The Cancer Imaging Archive (TCIA) \cite{clark2013cancer}, are applied in this study. We select the C-index as the evaluation metric. According to Supplementary Figure \ref{supfig:cindexsurvival}, we observe consistent performance improvements in PFMs equipped with ST information from STPath and protein expression information from \method {}, especially in the HNSC dataset. Thus, we have demonstrated that incorporating protein expression levels described by mIF not only improves the performance of unsupervised learning but also enhances the accuracy of prediction tasks. This provides further evidence for pathology learning supported by multimodal inputs and opens up new possibilities.

\begin{figure}[H]
    \centering
    \includegraphics[width=1.0\linewidth]{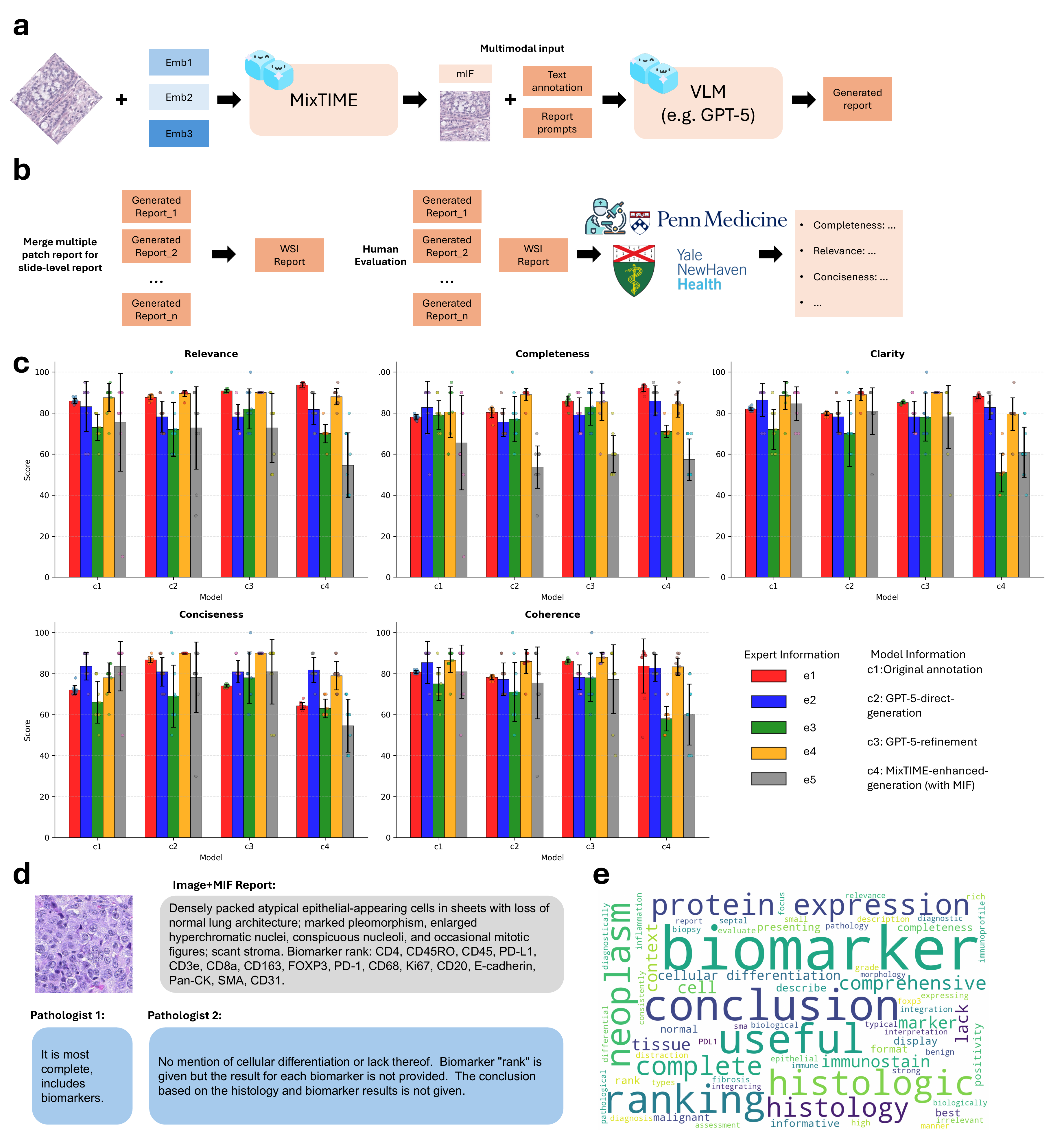}
    \caption{Pathology report generation with biomarkers. (a) Workflow of our pipeline for generating pathology reports from ROIs. (b) Workflow of WSI report generation and evaluation process with five pathologists from different institutes as well as five metrics covering different aspects. (c) Evaluation scores for reports from different methods by metrics. We annotate expert id and method information in the same figure (d) Case study for a pathology report with mIF information as well as comments from pathologists. (e) Word cloud of comments for \method{}-enhanced-generation reports made by pathologists.}
    \label{fig:report}
\end{figure}

\textbf{Generating medical reports based on multi-modal information.} 
A pathology report serves as a key carrier for describing the clinical and biological information associated with ROIs or WSIs. Thanks to advances in artificial intelligence, multimodal LLMs can also be used to generate pathology reports. By extracting features from images, these models can describe pathological structures and tissue characteristics, thereby providing guidance to doctors and patients for real-time diagnosis or condition assessment. However, current report-generation models mainly focus on image features, rather than other possibly important biomedical markers such as genes and proteins. Since most of this biological information can only be obtained through molecular testing, this information gap reduces the quality of the report. Here, we leverage the contributions of \method{} to predict protein expression levels using ROIs and generate medical reports with multimodal LLMs based on both image and biological signals. Our pipeline is shown in Figures \ref{fig:report} (a) and (b), where we extract the informative patches from a WSI and generate mIF as well as medical reports, with the help of \method{} and GPT-5. To investigate whether the information provided by \method{} helps improve report quality, we invited four pathologists from different institutions to evaluate the quality of the reports based on the 10 ROIs from one WSI (11 reports in total) across five criteria: Completeness, Relevance, Conciseness, Coherence, and Clarity. Details of the criteria can be found in the Methods section. The experiment is blinded to make a fair comparison.

The reports are provided with four different approaches: 1. Original annotation (human annotation with image features only); 2. GPT-5-direct-generation (prompting GPT-5 with report generation instructions and ROIs); 3. GPT-5-refinement (prompting GPT-5 with original annotation and ROIs to refine the report); and 4. \method{}-enhanced-generation (prompting GPT-5 with the rank of biomarker expressions, original annotation, and ROIs). Figure \ref{fig:report} (c) shows consistent variation across both models and experts, indicating that the evaluation is not driven by a single physician’s preference but reflects diverse expert judgments. Across the five evaluation dimensions, c4 (the \method{}-enhanced generation model) generally achieves competitive or strong performance, particularly in Relevance and Completeness, where its scores are often close to or above those of the other reports. Although c4 does not uniformly dominate every category (for example, its performance appears more variable in Clarity and lower in Conciseness), the results suggest that incorporating mIF-enhanced generation helps produce responses that are more aligned with some experts' expectations in several clinically important aspects. If we disregard inter-pathologist variability, as shown in Supplementary Figure \ref{supfig:addinforeport} (a), the integration of mIF yields scores comparable to those of other methods only in the Completeness category and does not outperform these models in other dimensions. Supplementary Figure \ref{supfig:addinforeport} (b) with statistical comparison also shows that introducing mIF information will reduce the Conciseness as we have more context, and thus the usage of this enhancement should be carefully evaluated before conducting. It is worth noting that reports generated directly by GPT-5 did not score significantly lower than the original annotations in manual evaluations; therefore, our approach of using GPT-5 to assist in report generation has been accepted by pathologists.

The visible spread among expert scores further highlights the importance of using multiple physicians rather than relying on a single evaluator: different experts emphasize different qualities of the responses, leading to meaningful variance within each model-category pair. As shown in Supplementary Figure \ref{supfig:addinforeport} (c), expert 5 (e5) is more reserved and generally does not prefer giving high scores for all models, and expert 2 (e2) is more optimistic and assigns high scores in general. This expert diversity strengthens the reliability of the evaluation by capturing heterogeneity in clinical interpretation, while the strong performance of c4 supports the contribution of \method{}-enhanced generation as a useful strategy for improving response quality beyond direct generation and refinement alone.

To account for differences in pathologists’ preferences and training backgrounds, and to further evaluate the contribution of biomarkers to report generation, we invited the pathologists participating in the study to attempt to describe the characteristics of reports from various sources. An example of reports from \method{}-enhanced-generation is shown in Figure \ref{fig:report} (d). From this ROI, we can see that the generated report includes organizational information such as lung architecture, as well as rankings of biomarker expression. Pathologist 1 acknowledges the introduction of biomarkers, while Pathologist 2 believes that the conclusion based on the histology and biomarker cannot be directly derived without exact expression levels. Thus, we can see that different pathologists have varying degrees of acceptance regarding the inclusion of molecular information; reports containing expression data are better suited for physicians with a background in molecular testing. Pathologists' evaluations of the report also share similar characteristics. Figure \ref{fig:report} (e) shows the word cloud of descriptions provided by pathologists for the reports generated by this method, and most of them identify the uniqueness of this type of report. As a preliminary attempt to explore the integration of molecular and image data, this study uses the rank of expression levels to characterize the aforementioned information. Overall, future research may need to consider describing mIF intensity in as much detail as possible, and the introduction of biomarker expression levels might be helpful for having a more comprehensive report in a shorter time.

\begin{figure}[H]
    \centering
    \includegraphics[width=1.0\linewidth]{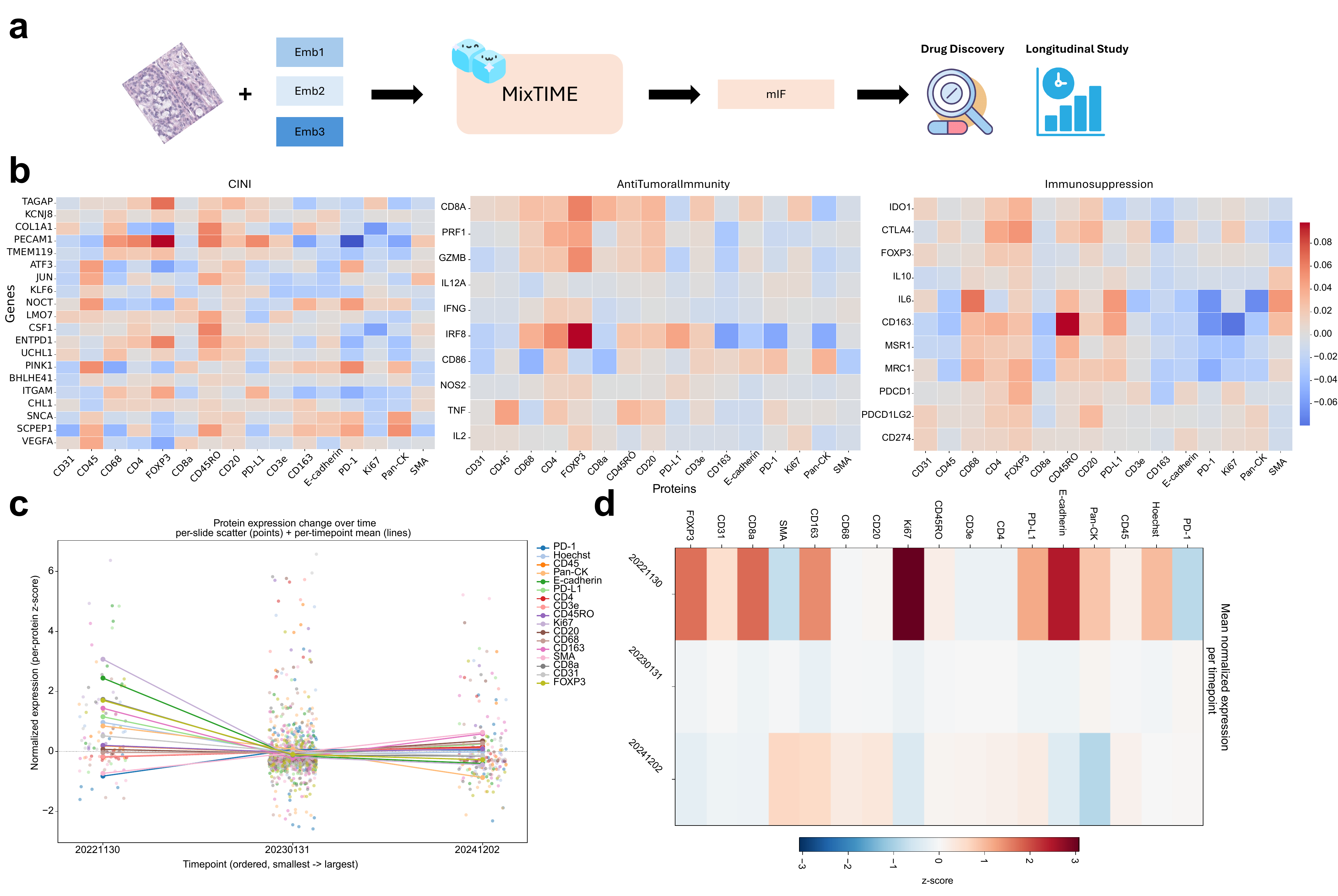}
    \caption{Applications of \method{} for disease analysis based on time series data. (a) Workflow of our pipeline for inferring mIF based on ROI and WSI. (b) Correlation heatmap between gene expression profiles and predicted mIF of proteins across three different sets of biomarker genes to investigate immune suppression. (c) Predicted mIF changes based on multiple slides from different time points in the investigated patients. (d) Averaged mIF expression heatmap across time points and proteins. For (c) and (d), the unit is a day.}
    \label{fig:ttanalysis}
\end{figure}

\textbf{Predicting protein expression on time shots uncovers hidden biomarkers for disease diagnosis and treatment.} \method{} not only can improve the accuracy of protein expression prediction and related phenotypes, but also enhances grounded discoveries for diagnosis-related biomarkers as well as treatment design, and thus \method{} serves as a potential tool for accelerating drug discovery focusing on specific tumor types. Our pipeline used in this section is illustrated in Figure \ref{fig:ttanalysis} (a).

Our first contribution is to predict proteins that have a strong correlation with genes serving as immunology signatures. Here, we focus on cutaneous squamous cell carcinoma samples to investigate gene signatures from one study of Peri-neural invasion (PNI). PNI is known as a well-established poor prognostic factor in multiple cancer types, and previous studies focused on analyzing the gene signatures of PNI, while lacking investigation of protein-related biomarkers to improve our understanding of drug resistance to anti-PD-1 therapy of squamous cell carcinoma \cite{baruch2025cancer}. To quantify the impact of proteins on immune responses, we extract HE images from all samples sequenced with ST \cite{baruch2025cancer} and use our method to predict their mIF intensity. For each patch, we calculate the correlation between the predicted mIF intensity of the 16 proteins and gene expression levels. According to Figure \ref{fig:ttanalysis} (b), we quantify the correlation between proteins and genes from three different categories: Cancer-Induced Nerve Injury (CINI), Anti-Tumoral-Immunity (AntiTumoralImmunity), and Immunosuppression. Interestingly, FOXP3 has been shown to have a strong expression correlation between one marker from both CINI and AntiTumoralImmunity. The strong correlation between FOXP3 (a regulatory T cell marker \cite{rudensky2011regulatory}) and PECAM1 (an endothelial/vascular marker \cite{privratsky2014pecam}) most likely reflects tumor microenvironment structure rather than co-expression in the same cells. In tumors, increased vascularization (high PECAM1) facilitates immune cell trafficking, leading to greater infiltration of multiple immune populations, including Tregs. Tregs also tend to accumulate near blood vessels due to endothelial signaling and perivascular niches, reinforcing this association. Moreover, the Immunosuppression gene signature IRF8, marks activated dendritic cells and macrophages that drive antigen presentation and interferon signaling, which promotes T cell recruitment—including both effector CD8 T cells and Tregs. At the same time, Tregs are often recruited or expanded in response to inflammation to limit excessive immune activation, and they can directly interact with IRF8-positive antigen-presenting cells (e.g., via CTLA4-mediated modulation \cite{qi2009differential}). Thus, FOXP3–IRF8 correlation reflects a balanced immune state where activation (IRF8-positive APCs) and suppression (FOXP3-positive Tregs) coexist. Based on our analysis, Treg cells play an important role in inhibiting the activity of tumor cells, and the interaction between FOXP3 and other related genes might help us develop better treatment plans.  

For gene signatures related to Immunosuppression, CD163 marks M2-like tumor-associated macrophages (TAMs) that produce IL-10/TGF-$\beta$ \cite{kwiecien2019cd163} and drive suppressive, pro-tumor signaling. CD45RA is expressed on naive T cells \cite{machura2008expression}, indicating a less activated or early-differentiation immune compartment. In suppressive tumors, M2 TAMs can (i) recruit/retain naive or non-effector T cells via chemokines, (ii) inhibit their activation and differentiation, and (iii) skew antigen presentation toward tolerance. The result is a tissue state where CD163-positive macrophage abundance co-occurs with a higher fraction of CD45RA-positive T cells, yielding a positive correlation. Therefore, the interaction between CD45RA and CD163 might be important to further understand the reduction of immune system effectiveness in the tumor microenvironment. 

In addition to the contributions mentioned above, \method{} can also help analyze changes in protein signatures across different time points and disease states, thereby facilitating the identification of pathological or therapeutic information at the molecular level. Here we leverage a set of whole slide images from one patient stored in Harvard Medical School; these slides have three different time points with paired annotations of electronic health records (EHRs) \ref{zhang2026multimodal}. By combining \method{} with TRIDENT \cite{zhang2025accelerating}, we can predict changes in multiple mIFs over time, thereby inferring correlations between protein expression levels and EHR data, and thus providing additional insights for disease diagnosis and treatment. 

Figures \ref{fig:ttanalysis} (c) and (d) illustrate longitudinal changes in the patient’s tumor microenvironment across three clinically meaningful time points (days). At the initial 2022 biopsy, when a 9 mm adenocarcinoma in the right middle lobe was confirmed, the protein profile showed relatively elevated epithelial/tumor-associated markers, including E-cadherin, Pan-CK, Ki67, CD8a, CD163, PD-L1, and FOXP3, suggesting an active tumor-associated epithelial and immune context. By 2023/1/31, after robotic right middle lobectomy with extensive pathological sampling and tumor-negative lymph nodes, most markers converged toward baseline, consistent with surgical removal of the malignant lesion and lack of nodal involvement. In 2024/12, when a new growing right upper lobe nodule raised concern for malignancy and the patient underwent robotic VATS extensive pneumolysis \cite{flores2008video}, RUL posterior segmentectomy, and MLND, the profile shifted again: several immune and stromal markers, including SMA, CD68, CD20, CD4, CD3e, and CD163, increased modestly, while epithelial markers such as Pan-CK and E-cadherin were not broadly elevated compared with the initial biopsy. Together, the scatter/line plot and heatmap suggest that the strongest tumor-associated epithelial and proliferative signal was present at the diagnostic biopsy, was largely reduced after definitive lobectomy, and was followed by a later immune/stromal remodeling pattern at the time of the new RUL lesion. Therefore, the predicted expression levels of these biomarkers can also be used as a basis for assessing disease severity.

We also provide an additional analysis based on another patients with slides from four different time points, and the changes of mIF strength are summarized in Supplementary Figures \ref{supfig:another example} (a) and (b). Across the longitudinal MIF profiles and EHR timeline, the most informative comparison is between the 2023/4/27 definitive resection and the 2024/4 repeat thoracic surgery. The 2023/4/27 specimen, pathologically confirmed as sarcomatoid malignant mesothelioma with extensive invasion of chest wall soft tissue, T2–T4 vertebrae, ribs, pleura, and lung, shows a broad increase in multiple protein signals compared with the initial biopsy, including immune and stromal-associated markers such as CD45, CD68, CD20, CD4/CD8a, CD31, and mesenchymal/tumor-context markers, consistent with a highly invasive tumor microenvironment accompanied by active inflammatory and stromal remodeling. In contrast, the 2024/4 recurrent/residual chest wall lesion, although still confirmed as sarcomatoid mesothelioma invading soft tissue and ribs, shows a more moderated MIF profile, with most markers closer to baseline and less broadly elevated than the 2023/4/27 resection, suggesting that the recurrent lesion after prior surgery and radiation may represent a more localized residual/recurrent tumor focus rather than the same degree of extensive, highly inflamed disease seen at the initial definitive resection. The negative margins in 2024 further support that this lesion was surgically contained, while the persistent mesothelioma diagnosis indicates ongoing local disease biology despite prior treatment. Overall, the paired MIF and EHR evidence suggest that 2023/4/27 captured the peak of aggressive, extensively invasive sarcomatoid mesothelioma with strong tumor–immune–stromal activation, whereas 2024/4 reflects locally recurrent/residual mesothelioma with comparatively reduced but persistent pathological activity. Again, \method{} successfully captures the changes at biomarker levels, and thus provide more information for the tumor microenvironment understanding.

\section{Discussion}
A central question in precision oncology is whether routinely available H\&E pathology images can be transformed into richer molecular representations of the tumor immune microenvironment. Multiplex immunofluorescence provides spatially resolved protein measurements that are highly informative for immune profiling, treatment response, and disease progression, but it remains costly, technically demanding, and difficult to scale across large clinical cohorts. Existing computational approaches have shown that mIF signals can be inferred from pathology images, yet most methods rely primarily on a single image modality and do not fully exploit complementary knowledge from image-text pathology models, spatial transcriptomic predictors, or previously trained mIF models. In this study, we address this gap by introducing \method{}, a multimodal mixture-of-expert pathology foundation model for high-resolution mIF prediction and tumor microenvironment analysis.

Therefore, we propose \method{} and make four major contributions to the modeling of multimodal pathology and precision oncology. Methodologically, it introduces a mixture-of-expert framework that integrates pathology foundation models trained with image-only, image-text, image-transcriptomic, and image-mIF supervision, allowing the model to dynamically weight complementary modalities while using a loss function that preserves both biomarker intensity and spatial distribution. Biologically and clinically, \method{} transforms H\&E images into predicted protein maps that serve as an additional molecular modality, improving mIF prediction, spatial domain identification, and survival prediction across multiple benchmarks. The model also enables biomarker-enhanced pathology report generation, where predicted mIF information improves clinically important aspects such as relevance and completeness, while also revealing differences in how pathologists interpret biomarker-enriched reports. Finally, \method{} supports longitudinal and treatment-related tumor microenvironment analysis by linking histology-derived protein profiles with immune gene programs, clinical trajectories, drug-resistance signatures, and disease-state changes, suggesting its potential as a scalable tool for molecular-informed pathology analysis and hypothesis generation.

This study also has several limitations. First, although \method{} improves mIF prediction across multiple datasets, prediction accuracy remains limited for some low-expression or noisy biomarkers. This likely reflects both biological difficulty and technical variation in mIF measurement, including staining quality, signal sparsity, batch effects, and tissue heterogeneity. Second, the current model is trained and evaluated using available public datasets, and additional validation across cancer types, institutions, scanners, staining protocols, and patient populations is needed before broad clinical deployment. Third, the report-generation experiment is preliminary, with a limited number of ROIs and expert evaluators. Although the blinded evaluation provides useful evidence, larger studies are needed to determine how biomarker-enhanced reports affect results. Future work should extend \method{} in several directions to address these concerns.

In summary, \method{} demonstrates that multimodal pathology foundation models can transform H\&E images into high-resolution immune biomarker predictions and use these predictions to support spatial domain identification, survival modeling, report generation, drug-resistance analysis, and longitudinal disease tracking. By integrating complementary knowledge from multiple expert models, \method{} moves digital pathology closer to molecular-informed precision oncology. While further validation and calibration are required, this work suggests a practical route toward scalable, multimodal tumor microenvironment profiling from routinely available pathology images.

\section{Methods}
\textbf{Problem definition.} Considering a whole-slide image $I$, which can be decomposed into several ($n$) regions of interest (ROIs) or patches $I=(i_1, i_2,...,i_n)$, our objective is to train a model $\mathcal{M}$ to predict the patch-level mIF information $m_k = \mathcal{M}(i_k)$. We can merge all the patches' predictions as the final mIF information as the slide-level mIF $M = (m_1, m_2,...,m_n)$. The prediction outcomes are treated as a new modality to enhance the original image information.

\textbf{MOE design.} Our model contains four different experts (FMs) with different resolutions and inputs. They are: 1. image-based expert (UNIv2 \cite{chen2024towards} $\mathcal{M}_1$), which takes H\&E image as input and image embedding as output. 2. image-text-based expert (CONCHv1.5 \cite{lu2024visual} $\mathcal{M}_2$), which takes H\&E image as input and image embedding as output. CONCHv1.5 is trained with paired text and image information. 3. image-transcriptomic-based expert (STPath \cite{huang2025stpath} $\mathcal{M}_3$), which takes H\&E image as input and gene expression as output. 4. image-mIF expert (GigaTIME \cite{valanarasu2026multimodal}, $\mathcal{M}_0$), which takes H\&E image as input and mIF strength as output. This expert serves as the backbone encoder-decoder and provides spatial feature maps for expert fusion. The other three experts inject their knowledge into $\mathcal{M}_0$'s feature maps via cross-attention. 

To determine the best combination of experts, we introduce a learnable router which can learn the weights of different experts' contributions, and we use pre-extracted embeddings from $\mathcal{M}_1$, $\mathcal{M}_2$, and $\mathcal{M}_3$ (frozen), while $\mathcal{M}_0$ and the fusion modules are fine-tuned during training. The output of each expert is linked to the cross-attention method \cite{huang2019ccnet}. Therefore, if we merge the router and cross-attention as the same unit $\mathcal{R}$, then the inference step can be written as:

\begin{equation}
    \hat{m}_k = \mathcal{R}(\mathcal{M}_0(i_k), \mathcal{M}_1(i_k), \mathcal{M}_2(i_k), \mathcal{M}_3(i_k)).
\end{equation}

For the HEMIT dataset \cite{bian2024hemit}, our output channel number is three. For the ORION dataset \cite{lin2023high}, our output channel number is 17. Unless otherwise specified, all downstream tasks use models trained on the Orion dataset.

\textbf{Model training.} The optimizer is AdamW \cite{loshchilov2017decoupled}, and hyper-parameters are tuned to their optimal setting based on the validation dataset. We use a learning rate of $1 \times 10^{-4}$, weight decay of $1 \times 10^{-5}$, batch size of 16, and train for up to 100 epochs. The cross-attention modules use 4 heads, 8 tokens, and dropout of 0.1. We set $\lambda_{ch} = \lambda_{px} = 1.0$. To learn a better representation of mIF, we leverage a loss function based on the mixture of negative Pearson Correlation Coefficients (PCCs) and Smooth L1 loss, which can help us learn the similarity of both tendency and magnitude. Assuming the model output is $\hat{m}_k \in \mathbb{R}^{C \times H \times W}$, and the observed mIF is $m_k$. We reshape both to $\mathbb{R}^{C \times N}$ where $N = B \times H \times W$. Our loss function is defined as:
\begin{equation}
\begin{aligned}
\mathcal{L}_{channel} &= 1 - \frac{1}{N}\sum_{j=1}^{N}\text{Pearson}(m_{(:,j)}, \hat{m}_{(:,j)}), \\
\mathcal{L}_{pixel} &= 1 - \frac{1}{C}\sum_{i=1}^{C}\text{Pearson}(m_{(i,:)}, \hat{m}_{(i,:)}), \\
\mathcal{L}_{final} &= \lambda_{ch}\mathcal{L}_{channel} + \lambda_{px}\mathcal{L}_{pixel} + \mathcal{L}_{SmoothL1}.
\end{aligned}
\end{equation}
where $C$ is the number of protein channels and $N$ is the total number of spatial positions. $\mathcal{L}_{channel}$ encourages consistent cross-channel predictions at each spatial location, while $\mathcal{L}_{pixel}$ encourages correct spatial distribution for each protein channel.

\textbf{Spatial domain identification.} Spatial domains are characterized based on biomolecular features from the patch-level image representations. Here we collect paired patches and domain labels, process the patches with different models, and produce embeddings. The embeddings from these methods are used to represent these patches and are evaluated by running the Leiden algorithm based on different resolutions (from 0.2 to 2.0). The Leiden algorithm can produce clustering labels based on an unsupervised learning approach, and we can evaluate the similarity between clustering labels and observed labels to evaluate model performance.

\textbf{Gene mutation and survival prediction.} Here we extract the prediction outcomes from \method{} as a new set of biomarkers, and build a multi-instance learning framework to integrate image features, embeddings from PFMs, and biomarkers from STPath and \method{} to predict slide-level annotation, including gene mutation and survival information. The patches are extracted based on TRIDET \cite{zhang2025standardizing, vaidya2025molecular}. 

\textbf{Medical report generation.} Here we test the quality of AI-generated medical reports with the help of information from different resources. We utilize GPT-5 \cite{singh2025openai} as the base generator, and control the input from image patches, textual annotation, and biomarker information predicted by \method{} with different combinations to generate medical reports based on predefined prompts. To enable the report generator to recognize molecular information, we normalize the protein expression levels and sort them from highest to lowest, then feed them into the model as a single sentence. Previous studies show that AI models specifically designed for pathology report generation generally cannot surpass frontier VLMs, and thus GPT-5 is a reasonable base model. We have three well-trained physicians to help us evaluate the quality of reports.

\textbf{Drug resistance analysis.} Here, we collect the image and transcriptomic information from a public ST dataset with drug resistance information, which includes 11 patients as well as genes from three groups: CINI, Anti-Tumoral-Immunity, and Immunosuppression. We predict the mIF strength for these datasets, and compute the correlation between different proteins and genes to identify patterns that show the difference between drug-sensitive and drug-resistant groups.

\textbf{Time-relevant biomarker prediction.} A key contribution of \method{} for this paper is the ability to predict protein expression levels over time, thereby identifying potential diagnostic markers and drug targets. Here, we collect ten patients with more than 100 visits per person, and each visit corresponds to one WSI. We extract the important patches and compute the average mIF strengths across different proteins, and analyze the expression changes as the patient's conditions become better or worse.

\textbf{Evaluations.} For mIF prediction, we refer to the setting from GigaTIME \cite{valanarasu2026multimodal} and utilize pixel-level Pearson Correlation Coefficients (PCCs) and Spearman Correlation Coefficients (SCCs) between observed mIFs and predicted mIFs as our main metrics. Both PCCs and SCCs range from -1 to 1, and higher scores represent a better method. For spatial domain clustering, we utilize Adjusted Rand Index (ARI), Adjusted Mutual Information (AMI), Homogeneity (HOMO), and Normalized Mutual Information (NMI) (four metrics in total) \cite{pedregosa2011scikit} for benchmarking different methods. All scores range from 0 to 1, and higher scores represent a better method. For gene mutation prediction, we utilize AUROC to evaluate. For survival prediction, we utilize C-index (AUROC in survival prediction) to evaluate \cite{pedregosa2011scikit}. All scores range from 0 to 1, and higher scores represent a better method.

To evaluate the quality of medical reports from different models, we provide grading criteria for physicians to consider and assign scores. The grading criteria contain five different dimensions:

\begin{itemize}
    \item Completeness: Definition: Measures whether the output fully addresses all required aspects of the task or question. Evaluation Criteria: Covers all key components requested; Does not omit critical steps, assumptions, or conclusions; Includes necessary context, explanations, or examples (when required).

\item Relevance: Relevance evaluates how well the model output aligns with the user’s intent and stays focused on the given task. A high score reflects content that directly responds to the prompt and avoids unnecessary or unrelated information. Responses that include tangential, off-topic, or distracting material should receive lower scores, especially if such content detracts from addressing the core request.

\item Conciseness: Conciseness assesses how efficiently the model communicates information without unnecessary verbosity. A high score indicates that the response is succinct, avoids redundancy, and includes only the level of detail appropriate for the task. Scores should be reduced for outputs that are overly long, repetitive, or padded with filler content, particularly when verbosity obscures the main message.

\item Coherence: Coherence measures the logical organization and flow of the response. High-scoring outputs present ideas in a clear, well-structured manner, with smooth transitions and consistent reasoning throughout. Lower scores should be given when the response is disorganized, contains abrupt shifts, internal contradictions, or a progression of ideas that is difficult to follow.

\item Clarity: Clarity evaluates how easily the response can be understood by the intended audience. A high score reflects precise language, well-formed sentences, and unambiguous explanations, with technical terms defined when necessary. Responses that are vague, confusing, or difficult to interpret should receive lower scores, especially if the lack of clarity hinders comprehension of the main points.
\end{itemize}

All scores range from 0 to 100, and higher scores represent a better method.

\textbf{Baselines.} For mIF prediction, we consider GigaTIME, MIPHEI-VIT \cite{balezo2026miphei}, ROSIE \cite{wu2025rosie}, and HEX \cite{li2026ai}. GigaTIME is a U-Net-based deep learning network pretrained with large-scale H\&E images and pixel-level mIF strength pairs. MIPHEI-VIT is a framework based on Visual Transformer (VIT) for predicting mIF strength at the pixel level. Its loss function also considers the variance of the protein expression. ROSIE is based on ConvNet and pretrained with large-scale patch-level HE images and mIF strength pairs from the CODEX datasets. HEX is also a patch-level mIF predictor based on deep neural networks. 

For spatial domain clustering, we consider both pre-trained PFMs such as UNIv2 \cite{chen2024towards}, and GigaPath \cite{xu2024whole}, but also spatial transcriptomic predictors, including TRIPLEX \cite{chung2024accurate}, BLEEP \cite{xie2023spatially}, and STPath \cite{huang2025stpath}. TRIPLEX integrates the image features from different resolutions to predict spatial transcriptomics, while BLEEP works as a retrieval-based deep learning model to predict spatial transcriptomics based on neighbor searching. The clustering results from UNI-based and GigaPath-based models are referred to as STPath.

For gene mutation and survival prediction, we consider both pre-trained PFMs, including UNIv2 and GigaPath, as well as the combination of these models' outputs with biomarkers predicted by STPath.

For medical report generation, we compare the medical reports based on three different resources (image only, human-annotated textual report, and image and biomarker information). We invite pathologists to grade the quality of these reports.

\textbf{Dataset Availability.} We have provided information on our training and testing datasets in Supplementary File 1, which are publicly available. The expertise information of selected pathologists for the evaluation of the report can be found in Supplementary File 2. For the TCGA datasets, please request access. For the WSIs used for time-relevant prediction, users need to receive approval from Harvard Medical School for access. Regarding the details of the pathologists' scoring, to protect personal privacy, we will not disclose them.

\textbf{Codes Availability.} Our codes can be found in: \url{https://github.com/HelloWorldLTY/MixTime}. Experiments with open-source models were run on a single NVIDIA H200 GPU, supported by NSF Access platform.

\textbf{Institutional Review Board (IRB) Approval.} This project has received approval from Yale IRB, with project number 2000039055. 

\section{Acknowledgments.}
We acknowledged one anonymous pathologist for assisting us in testing the report generation ability. This project is partly funded by the NSF Access Discovery program.

\section{Author contributions.}
T.L. designed this study. T.L. ran experiments with Z.W., Z.L., and T.D. P.H., L.C.C, K.T.E.C., and E.L.L.P. performed human evaluations. M.K. and J.C.K.L. organized the human evaluator team. All of the authors wrote and reviewed this manuscript. T.L. and W.J. jointly supervised this project.

\section{Competing interests.}
The authors declare no competing interests.

\bibliographystyle{unsrt}
\bibliography{sn-bibliography}

@article{nagarsheth2017chemokines,
  title={Chemokines in the cancer microenvironment and their relevance in cancer immunotherapy},
  author={Nagarsheth, Nisha and Wicha, Max S and Zou, Weiping},
  journal={Nature Reviews Immunology},
  volume={17},
  number={9},
  pages={559--572},
  year={2017},
  publisher={Nature Publishing Group UK London}
}

@article{wang2024immunotherapy,
  title={Immunotherapy and the ovarian cancer microenvironment: exploring potential strategies for enhanced treatment efficacy},
  author={Wang, Zhi-Bin and Zhang, Xiu and Fang, Chao and Liu, Xiao-Ting and Liao, Qian-Jin and Wu, Nayiyuan and Wang, Jing},
  journal={Immunology},
  volume={173},
  number={1},
  pages={14--32},
  year={2024},
  publisher={Wiley Online Library}
}

@article{bagaev2021conserved,
  title={Conserved pan-cancer microenvironment subtypes predict response to immunotherapy},
  author={Bagaev, Alexander and Kotlov, Nikita and Nomie, Krystle and Svekolkin, Viktor and Gafurov, Azamat and Isaeva, Olga and Osokin, Nikita and Kozlov, Ivan and Frenkel, Felix and Gancharova, Olga and others},
  journal={Cancer cell},
  volume={39},
  number={6},
  pages={845--865},
  year={2021},
  publisher={Elsevier}
}

@article{binnewies2018understanding,
  title={Understanding the tumor immune microenvironment (TIME) for effective therapy},
  author={Binnewies, Mikhail and Roberts, Edward W and Kersten, Kelly and Chan, Vincent and Fearon, Douglas F and Merad, Miriam and Coussens, Lisa M and Gabrilovich, Dmitry I and Ostrand-Rosenberg, Suzanne and Hedrick, Catherine C and others},
  journal={Nature medicine},
  volume={24},
  number={5},
  pages={541--550},
  year={2018},
  publisher={Nature Publishing Group}
}

@article{chen2017computer,
  title={Computer-aided prognosis on breast cancer with hematoxylin and eosin histopathology images: A review},
  author={Chen, Jia-Mei and Li, Yan and Xu, Jun and Gong, Lei and Wang, Lin-Wei and Liu, Wen-Lou and Liu, Juan},
  journal={Tumor Biology},
  volume={39},
  number={3},
  pages={1010428317694550},
  year={2017},
  publisher={SAGE Publications Sage UK: London, England}
}

@article{valanarasu2026multimodal,
  title={Multimodal AI generates virtual population for tumor microenvironment modeling},
  author={Valanarasu, Jeya Maria Jose and Xu, Hanwen and Usuyama, Naoto and Kim, Chanwoo and Wong, Cliff and Argaw, Peniel and Shimol, Racheli Ben and Crabtree, Angela and Matlock, Kevin and Bartlett, Alexandra Q and others},
  journal={Cell},
  volume={189},
  number={2},
  pages={386--400},
  year={2026},
  publisher={Elsevier}
}

@article{huang2025stpath,
  title={STPath: a generative foundation model for integrating spatial transcriptomics and whole-slide images},
  author={Huang, Tinglin and Liu, Tianyu and Babadi, Mehrtash and Ying, Rex and Jin, Wengong},
  journal={NPJ Digital Medicine},
  volume={8},
  number={1},
  pages={659},
  year={2025},
  publisher={Nature Publishing Group UK London}
}

@article{liu2026leveraging,
  title={Leveraging multi-modal foundation models for analysing spatial multi-omic and histopathology data},
  author={Liu, Tianyu and Huang, Tinglin and Ding, Tong and Wu, Hao and Humphrey, Peter and Perincheri, Sudhir and Schalper, Kurt and Ying, Rex and Xu, Hua and Zou, James and others},
  journal={Nature Biomedical Engineering},
  pages={1--18},
  year={2026},
  publisher={Nature Publishing Group UK London}
}

@article{chen2025visual,
  title={A visual--omics foundation model to bridge histopathology with spatial transcriptomics},
  author={Chen, Weiqing and Zhang, Pengzhi and Tran, Tu N and Xiao, Yiwei and Li, Shengyu and Shah, Vrutant V and Cheng, Hao and Brannan, Kristopher W and Youker, Keith and Lai, Li and others},
  journal={Nature Methods},
  volume={22},
  number={7},
  pages={1568--1582},
  year={2025},
  publisher={Nature Publishing Group US New York}
}

@article{lu2024visual,
  title={A visual-language foundation model for computational pathology},
  author={Lu, Ming Y and Chen, Bowen and Williamson, Drew FK and Chen, Richard J and Liang, Ivy and Ding, Tong and Jaume, Guillaume and Odintsov, Igor and Le, Long Phi and Gerber, Georg and others},
  journal={Nature medicine},
  volume={30},
  number={3},
  pages={863--874},
  year={2024},
  publisher={Nature Publishing Group US New York}
}

@article{odell2013immunofluorescence,
  title={Immunofluorescence techniques},
  author={Odell, Ian D and Cook, Deborah},
  journal={Journal of Investigative Dermatology},
  volume={133},
  number={1},
  pages={1--4},
  year={2013},
  publisher={Elsevier}
}

@article{balezo2026miphei,
  title={MIPHEI-ViT: Multiplex immunofluorescence prediction from H\&E images using ViT foundation models},
  author={Balezo, Guillaume and Trullo, Roger and Planas, Albert Pla and Decenci{\`e}re, Etienne and Walter, Thomas},
  journal={Computers in Biology and Medicine},
  volume={206},
  pages={111564},
  year={2026},
  publisher={Elsevier}
}

@article{wu2025rosie,
  title={ROSIE: AI generation of multiplex immunofluorescence staining from histopathology images},
  author={Wu, Eric and Bieniosek, Matthew and Wu, Zhenqin and Thakkar, Nitya and Charville, Gregory W and Makky, Ahmad and Sch{\"u}rch, Christian M and Huyghe, Jeroen R and Peters, Ulrike and Li, Christopher I and others},
  journal={Nature Communications},
  volume={16},
  number={1},
  pages={7633},
  year={2025},
  publisher={Nature Publishing Group UK London}
}

@article{li2026ai,
  title={AI-enabled virtual spatial proteomics from histopathology for interpretable biomarker discovery in lung cancer},
  author={Li, Zhe and Li, Yuchen and Xiang, Jinxi and Wang, Xiyue and Yang, Sen and Zhang, Xiaoming and Eweje, Feyisope and Chen, Yijiang and Luo, Xiangde and Li, Yuanyuan and others},
  journal={Nature Medicine},
  pages={1--14},
  year={2026},
  publisher={Nature Publishing Group US New York}
}

@article{chen2024towards,
  title={Towards a general-purpose foundation model for computational pathology},
  author={Chen, Richard J and Ding, Tong and Lu, Ming Y and Williamson, Drew FK and Jaume, Guillaume and Song, Andrew H and Chen, Bowen and Zhang, Andrew and Shao, Daniel and Shaban, Muhammad and others},
  journal={Nature medicine},
  volume={30},
  number={3},
  pages={850--862},
  year={2024},
  publisher={Nature Publishing Group US New York}
}

@inproceedings{huang2019ccnet,
  title={Ccnet: Criss-cross attention for semantic segmentation},
  author={Huang, Zilong and Wang, Xinggang and Huang, Lichao and Huang, Chang and Wei, Yunchao and Liu, Wenyu},
  booktitle={Proceedings of the IEEE/CVF international conference on computer vision},
  pages={603--612},
  year={2019}
}

@article{shazeer2017outrageously,
  title={Outrageously large neural networks: The sparsely-gated mixture-of-experts layer},
  author={Shazeer, Noam and Mirhoseini, Azalia and Maziarz, Krzysztof and Davis, Andy and Le, Quoc and Hinton, Geoffrey and Dean, Jeff},
  journal={arXiv preprint arXiv:1701.06538},
  year={2017}
}

@inproceedings{bian2024hemit,
  title={Hemit: H\&e to multiplex-immunohistochemistry image translation with dual-branch pix2pix generator},
  author={Bian, Chang and Phillips, Beth and Cootes, Tim and Fergie, Martin},
  booktitle={International Conference on Medical Image Computing and Computer-Assisted Intervention},
  pages={184--197},
  year={2024},
  organization={Springer}
}

@article{lin2023high,
  title={High-plex immunofluorescence imaging and traditional histology of the same tissue section for discovering image-based biomarkers},
  author={Lin, Jia-Ren and Chen, Yu-An and Campton, Daniel and Cooper, Jeremy and Coy, Shannon and Yapp, Clarence and Tefft, Juliann B and McCarty, Erin and Ligon, Keith L and Rodig, Scott J and others},
  journal={Nature cancer},
  volume={4},
  number={7},
  pages={1036--1052},
  year={2023},
  publisher={Nature Publishing Group US New York}
}

@article{zhang2025standardizing,
  title={Accelerating Data Processing and Benchmarking of AI Models for Pathology},
  author={Zhang, Andrew and Jaume, Guillaume and Vaidya, Anurag and Ding, Tong and Mahmood, Faisal},
  journal={arXiv preprint arXiv:2502.06750},
  year={2025}
}

@article{vaidya2025molecular,
  title={Molecular-driven Foundation Model for Oncologic Pathology},
  author={Vaidya, Anurag and Zhang, Andrew and Jaume, Guillaume and Song, Andrew H and Ding, Tong and Wagner, Sophia J and Lu, Ming Y and Doucet, Paul and Robertson, Harry and Almagro-Perez, Cristina and others},
  journal={arXiv preprint arXiv:2501.16652},
  year={2025}
}

@article{singh2025openai,
  title={Openai gpt-5 system card},
  author={Singh, Aaditya and Fry, Adam and Perelman, Adam and Tart, Adam and Ganesh, Adi and El-Kishky, Ahmed and McLaughlin, Aidan and Low, Aiden and Ostrow, AJ and Ananthram, Akhila and others},
  journal={arXiv preprint arXiv:2601.03267},
  year={2025}
}

@article{pedregosa2011scikit,
  title={Scikit-learn: Machine learning in Python},
  author={Pedregosa, Fabian and Varoquaux, Ga{\"e}l and Gramfort, Alexandre and Michel, Vincent and Thirion, Bertrand and Grisel, Olivier and Blondel, Mathieu and Prettenhofer, Peter and Weiss, Ron and Dubourg, Vincent and others},
  journal={the Journal of machine Learning research},
  volume={12},
  pages={2825--2830},
  year={2011},
  publisher={JMLR. org}
}

@article{xu2024whole,
  title={A whole-slide foundation model for digital pathology from real-world data},
  author={Xu, Hanwen and Usuyama, Naoto and Bagga, Jaspreet and Zhang, Sheng and Rao, Rajesh and Naumann, Tristan and Wong, Cliff and Gero, Zelalem and Gonz{\'a}lez, Javier and Gu, Yu and others},
  journal={Nature},
  volume={630},
  number={8015},
  pages={181--188},
  year={2024},
  publisher={Nature Publishing Group UK London}
}

@inproceedings{chung2024accurate,
  title={Accurate spatial gene expression prediction by integrating multi-resolution features},
  author={Chung, Youngmin and Ha, Ji Hun and Im, Kyeong Chan and Lee, Joo Sang},
  booktitle={Proceedings of the IEEE/CVF Conference on Computer Vision and Pattern Recognition},
  pages={11591--11600},
  year={2024}
}

@article{xie2023spatially,
  title={Spatially resolved gene expression prediction from histology images via bi-modal contrastive learning},
  author={Xie, Ronald and Pang, Kuan and Chung, Sai and Perciani, Catia and MacParland, Sonya and Wang, Bo and Bader, Gary},
  journal={Advances in Neural Information Processing Systems},
  volume={36},
  pages={70626--70637},
  year={2023}
}

@inbook{pytorchframework,
author = {Paszke, Adam and Gross, Sam and Massa, Francisco and Lerer, Adam and Bradbury, James and Chanan, Gregory and Killeen, Trevor and Lin, Zeming and Gimelshein, Natalia and Antiga, Luca and Desmaison, Alban and K\"{o}pf, Andreas and Yang, Edward and DeVito, Zach and Raison, Martin and Tejani, Alykhan and Chilamkurthy, Sasank and Steiner, Benoit and Fang, Lu and Bai, Junjie and Chintala, Soumith},
title = {PyTorch: an imperative style, high-performance deep learning library},
year = {2019},
publisher = {Curran Associates Inc.},
address = {Red Hook, NY, USA},
booktitle = {Proceedings of the 33rd International Conference on Neural Information Processing Systems},
articleno = {721},
numpages = {12}
}

@article{chen2024stimage,
  title={Stimage-1k4m: A histopathology image-gene expression dataset for spatial transcriptomics},
  author={Chen, Jiawen and Zhou, Muqing and Wu, Wenrong and Zhang, Jinwei and Li, Yun and Li, Didong},
  journal={Advances in neural information processing systems},
  volume={37},
  pages={35796--35823},
  year={2024}
}

@article{baruch2025cancer,
  title={Cancer-induced nerve injury promotes resistance to anti-PD-1 therapy},
  author={Baruch, Erez N and Gleber-Netto, Frederico O and Nagarajan, Priyadharsini and Rao, Xiayu and Akhter, Shamima and Eichwald, Tuany and Xie, Tongxin and Balood, Mohammad and Adewale, Adebayo and Naara, Shorook and others},
  journal={Nature},
  volume={646},
  number={8084},
  pages={462--473},
  year={2025},
  publisher={Nature Publishing Group UK London}
}

@article{rudensky2011regulatory,
  title={Regulatory T cells and Foxp3},
  author={Rudensky, Alexander Y},
  journal={Immunological reviews},
  volume={241},
  number={1},
  pages={260--268},
  year={2011},
  publisher={Wiley Online Library}
}

@article{privratsky2014pecam,
  title={PECAM-1: regulator of endothelial junctional integrity},
  author={Privratsky, Jamie R and Newman, Peter J},
  journal={Cell and tissue research},
  volume={355},
  number={3},
  pages={607--619},
  year={2014},
  publisher={Springer}
}

@article{qi2009differential,
  title={Differential expression of IRF8 in subsets of macrophages and dendritic cells and effects of IRF8 deficiency on splenic B cell and macrophage compartments},
  author={Qi, Chen-Feng and Li, Zhaoyang and Raffeld, Mark and Wang, Hongsheng and Kovalchuk, Alexander L and Morse III, Herbert C},
  journal={Immunologic research},
  volume={45},
  number={1},
  pages={62--74},
  year={2009},
  publisher={Springer}
}

@article{kwiecien2019cd163,
  title={CD163 and CCR7 as markers for macrophage polarisation in lung cancer microenvironment},
  author={Kwiecie{\'n}, Iwona and Polubiec-Kownacka, Ma{\l}gorzata and Dziedzic, Dariusz and Wo{\l}osz, Dominika and Rzepecki, Piotr and Domaga{\l}a-Kulawik, Joanna},
  journal={Central European Journal of Immunology},
  volume={44},
  number={4},
  pages={395--402},
  year={2019},
  publisher={Termedia}
}

@article{machura2008expression,
  title={Expression of naive/memory (CD45RA/CD45RO) markers by peripheral blood CD4+ and CD8+ T cells in children with asthma},
  author={Machura, Edyta and Mazur, Bogdan and Pieni{\k{a}}{\.z}ek, Wojciech and Karczewska, Krystyna},
  journal={Archivum immunologiae et therapiae experimentalis},
  volume={56},
  number={1},
  pages={55--62},
  year={2008},
  publisher={Springer}
}

@article{liu2025unicorn,
  title={UNICORN: Towards universal cellular expression prediction with a multi-task learning framework},
  author={Liu, Tianyu and Huang, Tinglin and Wang, Lijun and Lin, Yingxin and Ying, Rex and Zhao, Hongyu},
  journal={Nature Communications},
  volume={16},
  number={1},
  pages={9455},
  year={2025},
  publisher={Nature Publishing Group UK London}
}

@article{clark2013cancer,
  title={The Cancer Imaging Archive (TCIA): maintaining and operating a public information repository},
  author={Clark, Kenneth and Vendt, Bruce and Smith, Kirk and Freymann, John and Kirby, Justin and Koppel, Paul and Moore, Stephen and Phillips, Stanley and Maffitt, David and Pringle, Michael and others},
  journal={Journal of digital imaging},
  volume={26},
  number={6},
  pages={1045--1057},
  year={2013},
  publisher={Springer}
}

@article{loshchilov2017decoupled,
  title={Decoupled weight decay regularization},
  author={Loshchilov, Ilya and Hutter, Frank},
  journal={arXiv preprint arXiv:1711.05101},
  year={2017}
}

@article{zhang2025accelerating,
  title={Accelerating data processing and benchmarking of ai models for pathology},
  author={Zhang, Andrew and Jaume, Guillaume and Vaidya, Anurag and Ding, Tong and Mahmood, Faisal},
  journal={arXiv preprint arXiv:2502.06750},
  year={2025}
}

@article{flores2008video,
  title={Video-assisted thoracic surgery lobectomy (VATS), open thoracotomy, and the robot for lung cancer},
  author={Flores, Raja M and Alam, Naveed},
  journal={The Annals of thoracic surgery},
  volume={85},
  number={2},
  pages={S710--S715},
  year={2008},
  publisher={Elsevier}
}

@article{zhang2026multimodal,
  title={A multimodal and temporal foundation model for virtual patient representations at healthcare system scale},
  author={Zhang, Andrew and Ding, Tong and Wagner, Sophia J and Tian, Caiwei and Lu, Ming Y and Pettit, Rowland and Lewis, Joshua E and Misrahi, Alexandre and Mo, Dandan and Le, Long Phi and others},
  journal={arXiv preprint arXiv:2604.18570},
  year={2026}
}

\appendix
\counterwithin{figure}{section}
\renewcommand{\figurename}{Supplementary Fig.}
\renewcommand\thefigure{\arabic{figure}} 

\newpage

\section{Supplementary figures}
\begin{figure}[H]
    \centering
    \includegraphics[width=1\linewidth]{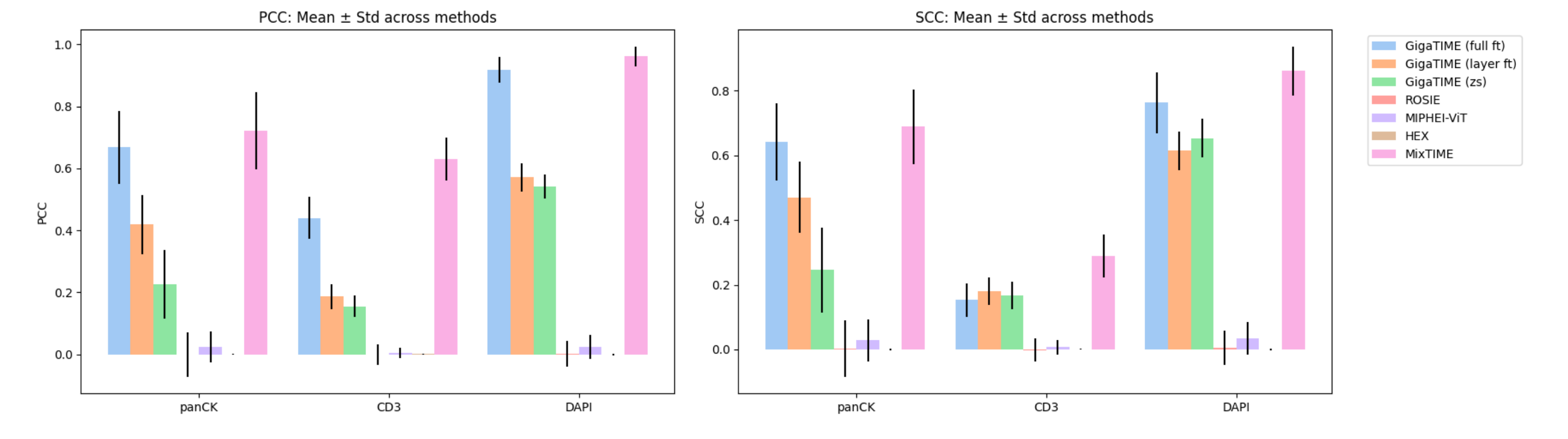}
    \caption{mIF prediction comparison across seven different methods.}
    \label{supfig:hemit result}
\end{figure}

\newpage

\begin{figure}[H]
    \centering
    \includegraphics[width=1\linewidth]{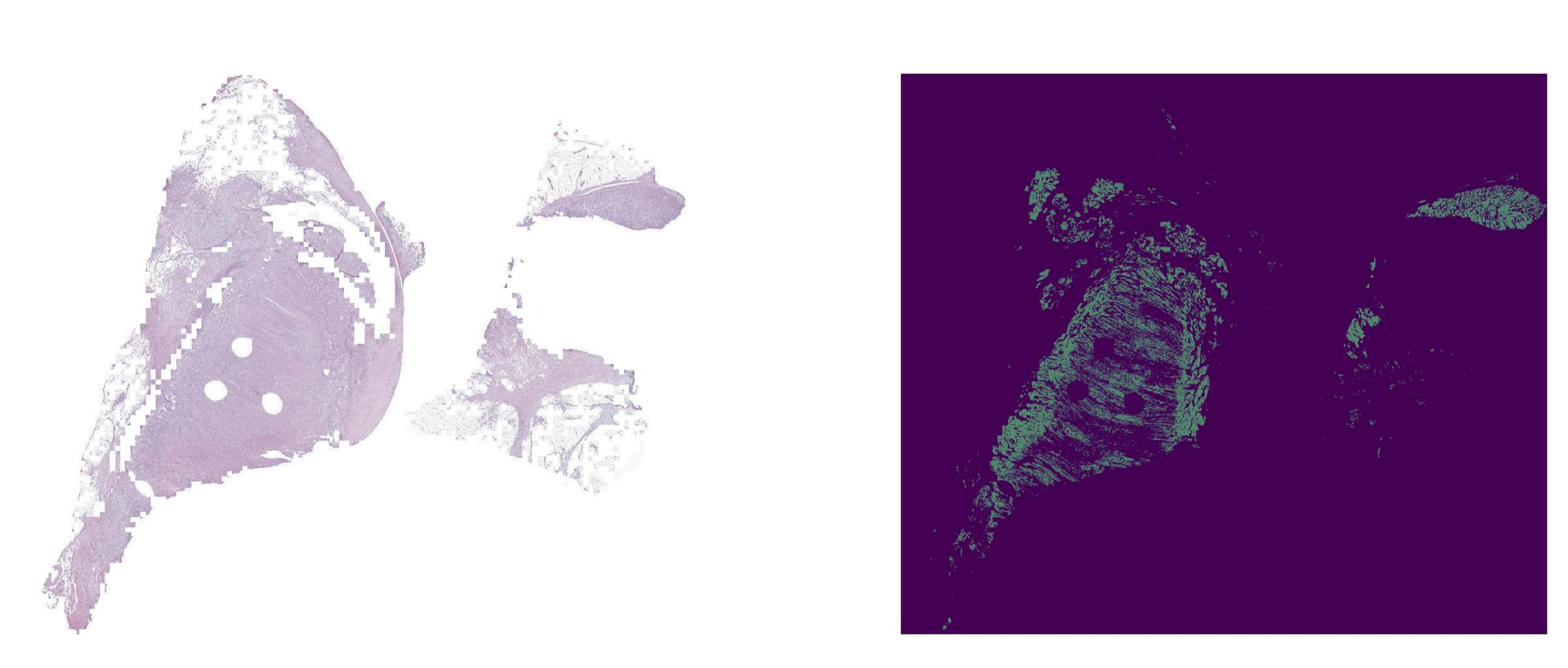}
    \caption{Visualization of whole-slide-image mIF prediction result base on protein PD1. The left panel represents slide image, and the right panel represents predicted protein expression levels.}
    \label{supfig:wholeslide}
\end{figure}

\newpage

\begin{figure}
    \centering
    \includegraphics[width=1\linewidth]{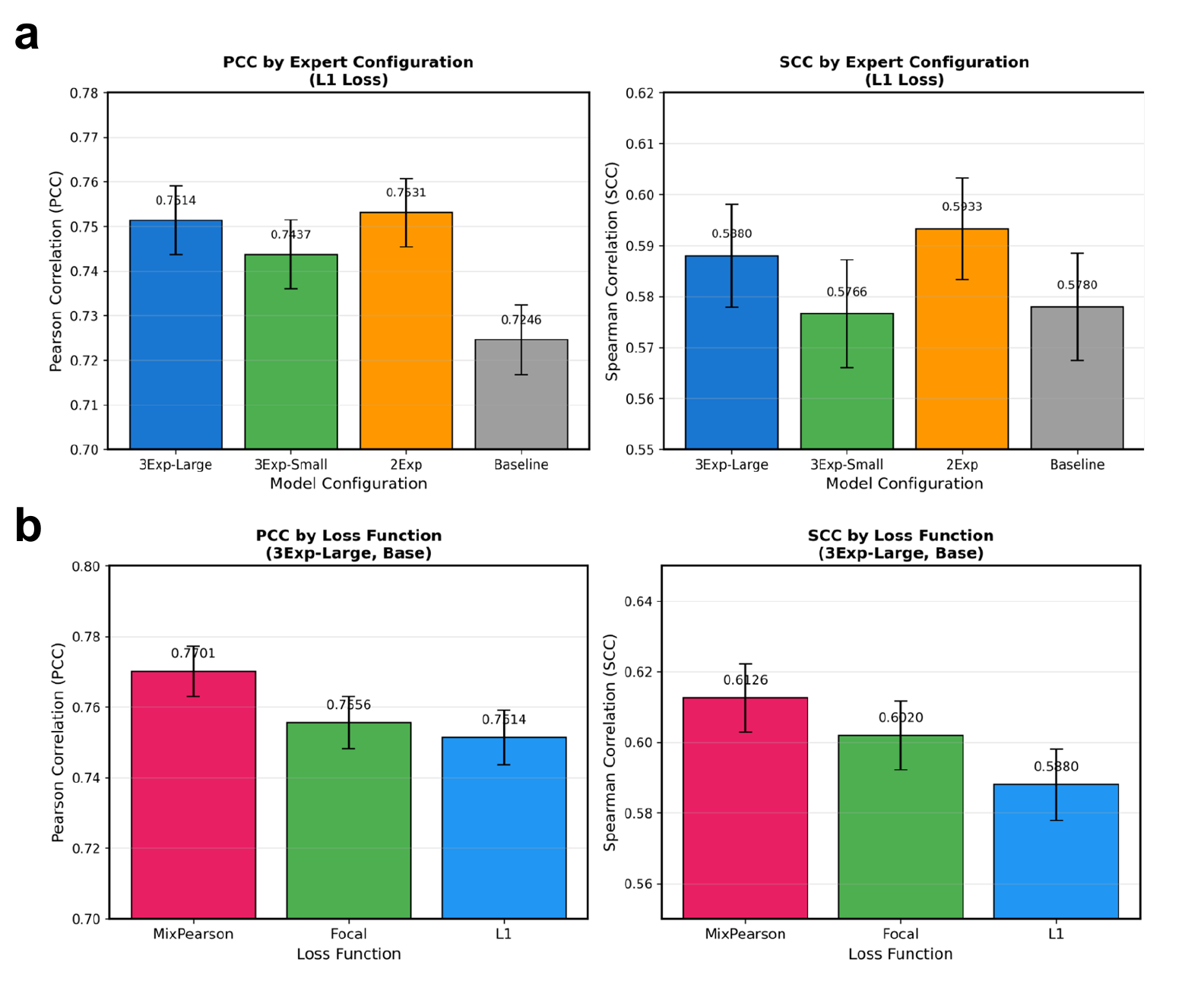}
    \caption{Visualization of ablation studies. (a) PCC and SCC by adjusting the choices of expert models. (b) PCC and SCC by adjusting the loss function components.}
    \label{supfig:abla_study}
\end{figure}

\begin{figure}
    \centering
    \includegraphics[width=1\linewidth]{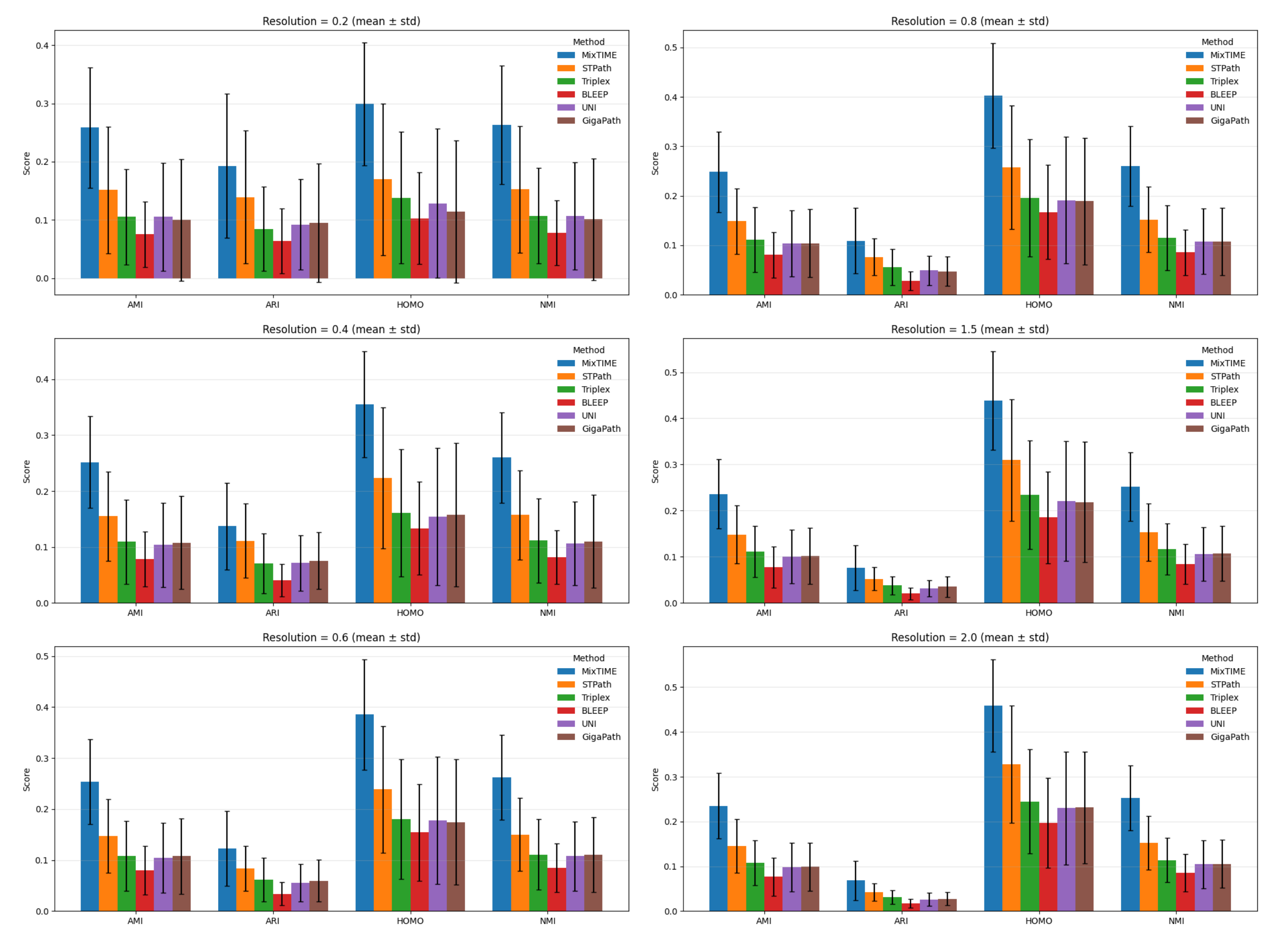}
    \caption{Clustering comparison across six different resolutions for dataset GSE213688.}
    \label{supfig:gse}
\end{figure}

\begin{figure}
    \centering
    \includegraphics[width=1\linewidth]{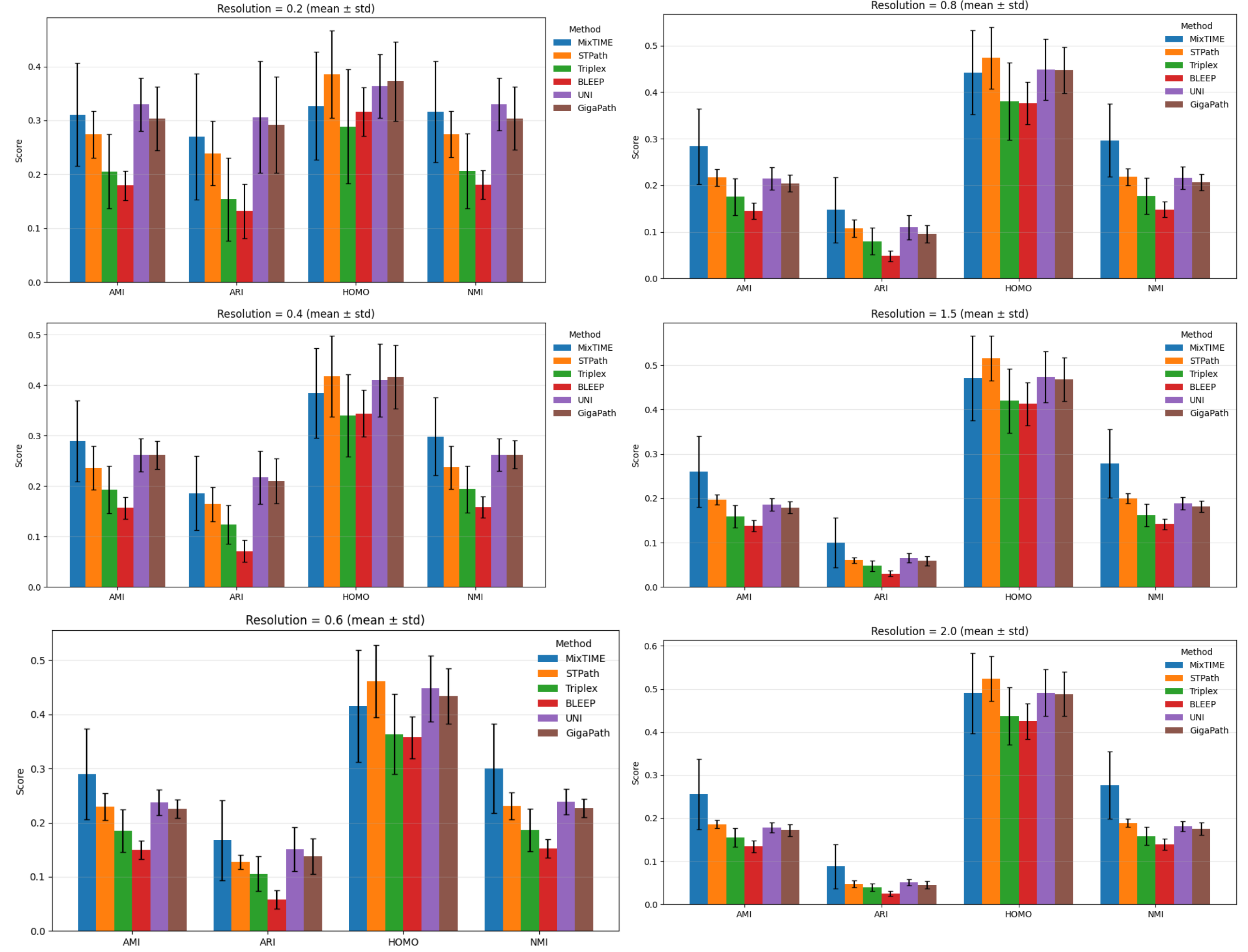}
    \caption{Clustering comparison across six different resolutions for dataset Erickson.}
    \label{supfig:erickson}
\end{figure}

\begin{figure}
    \centering
    \includegraphics[width=1\linewidth]{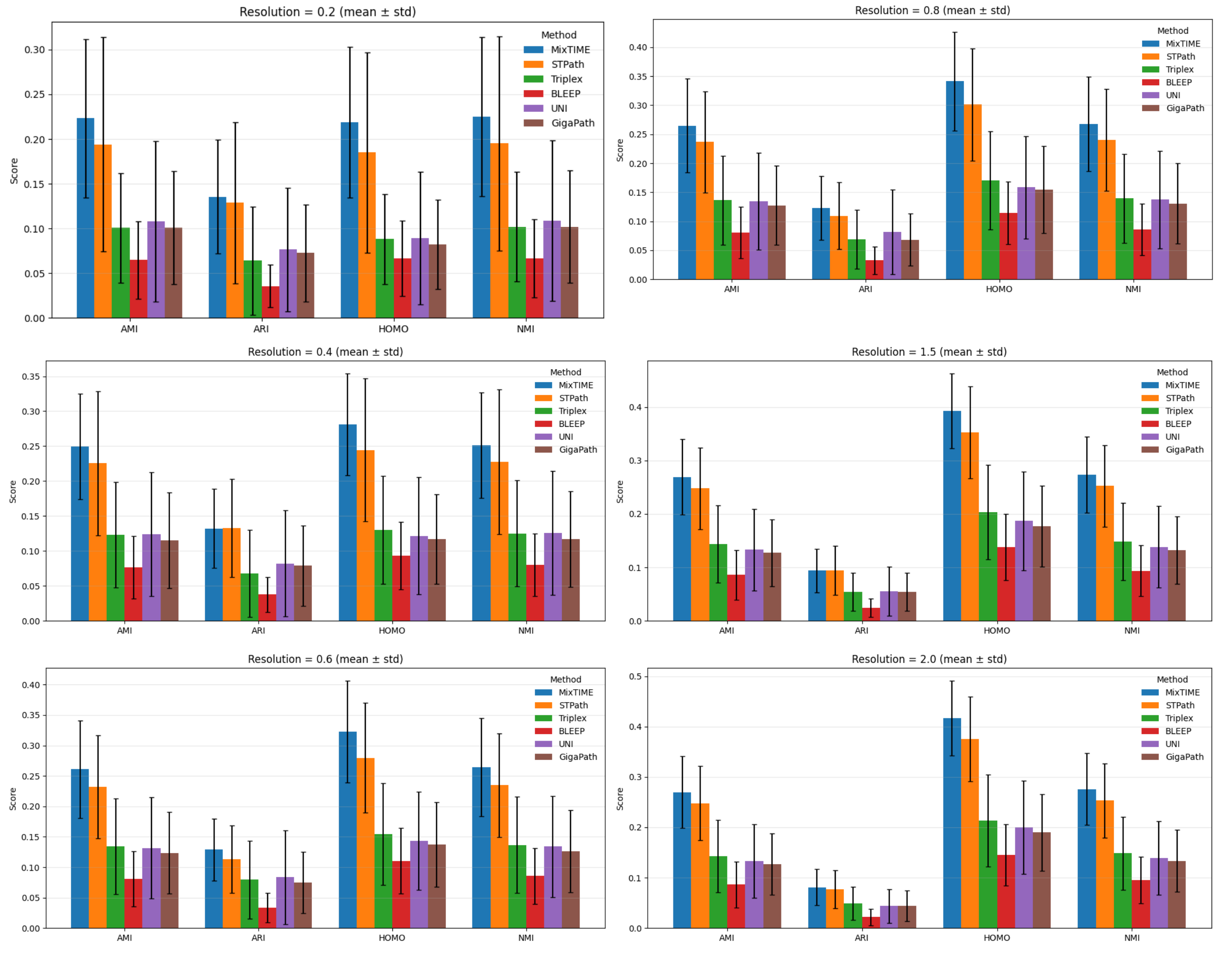}
    \caption{Clustering comparison across six different resolutions for dataset Brain.}
    \label{supfig:brain}
\end{figure}

\begin{figure}
    \centering
    \includegraphics[width=1\linewidth]{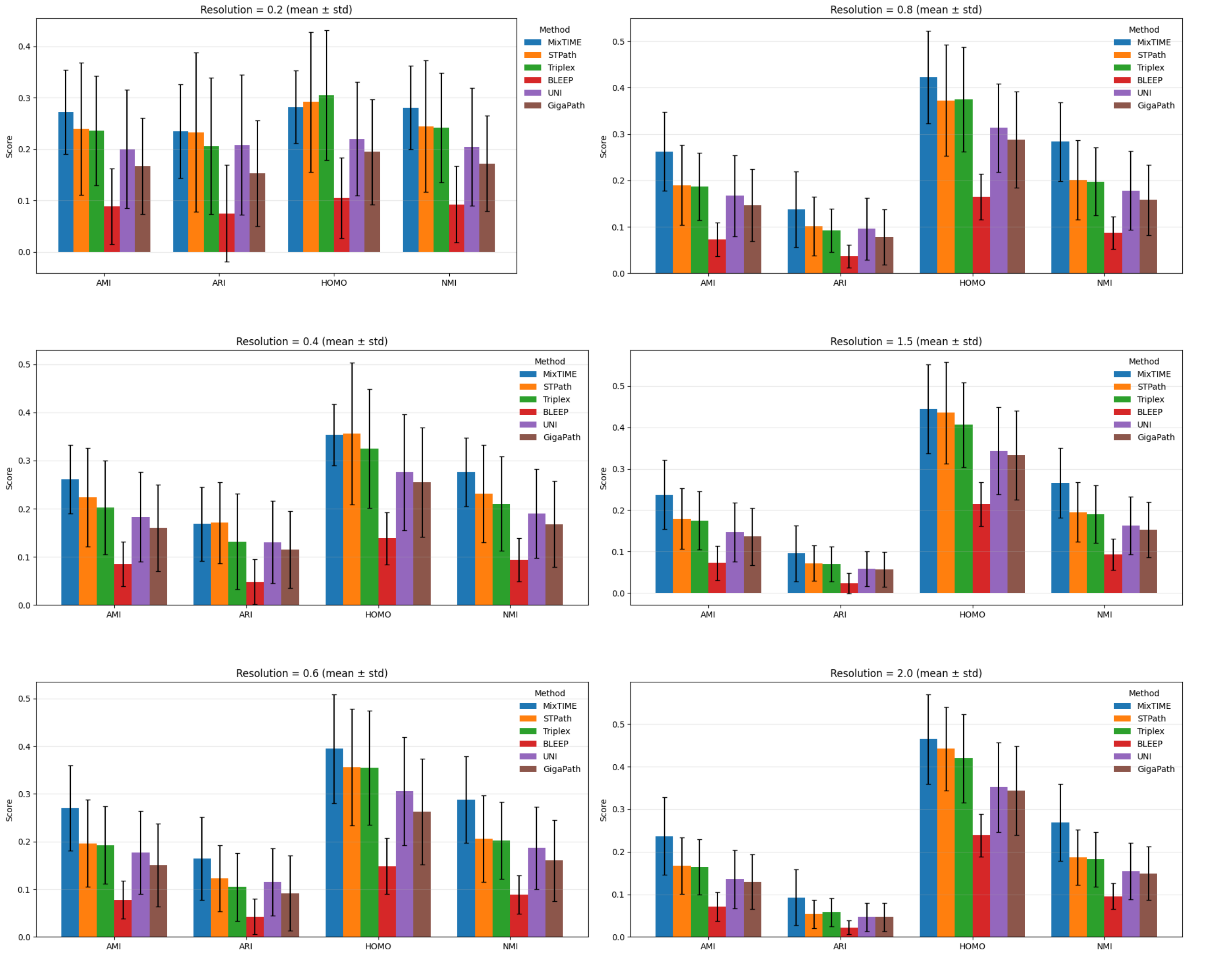}
    \caption{Clustering comparison across six different resolutions for dataset Andersson.}
    \label{supfig:andersion}
\end{figure}

\clearpage

\begin{figure}
    \centering
    \includegraphics[width=1\linewidth]{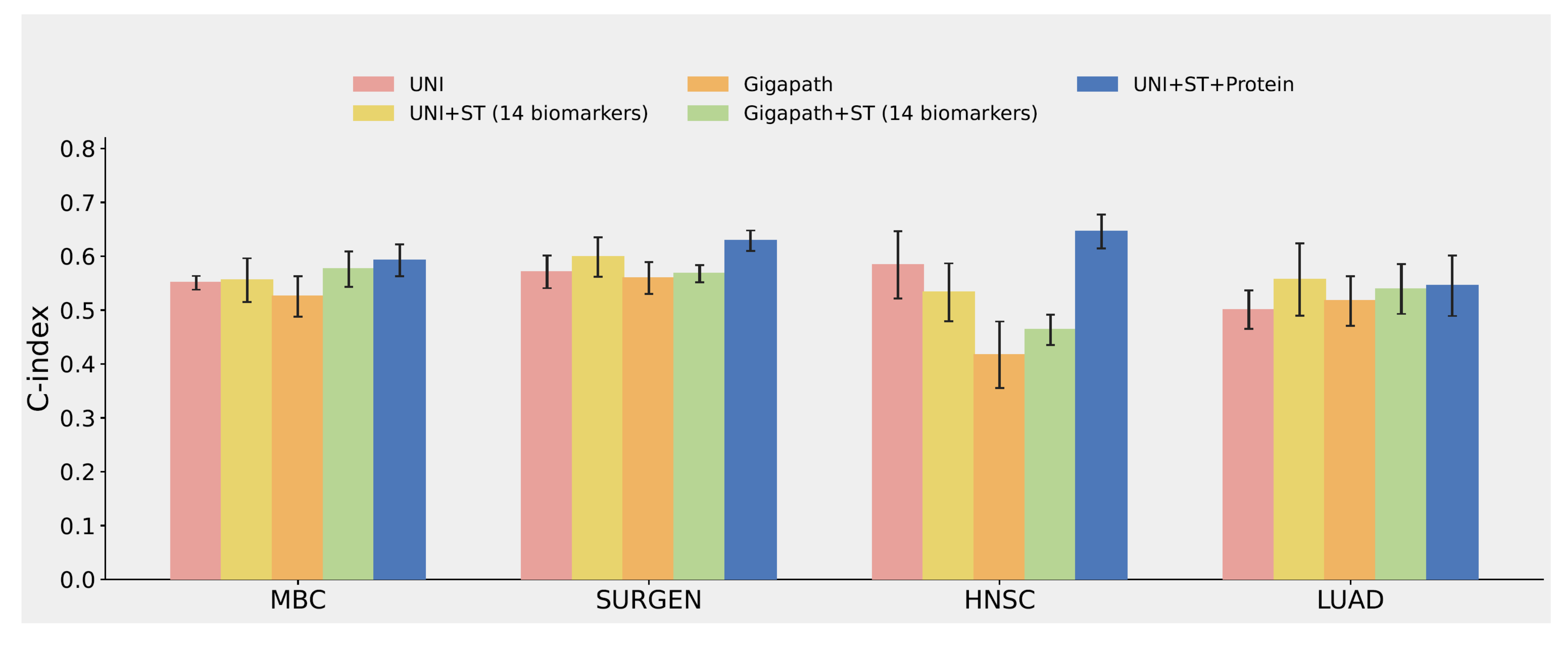}
    \caption{Results of survival prediction comparison across five different settings. Here 14 biomarkers represent 14 genes selected by STPath based on disease marker genes.}
    \label{supfig:cindexsurvival}
\end{figure}

\clearpage

\begin{figure}
    \centering
    \includegraphics[width=1\linewidth]{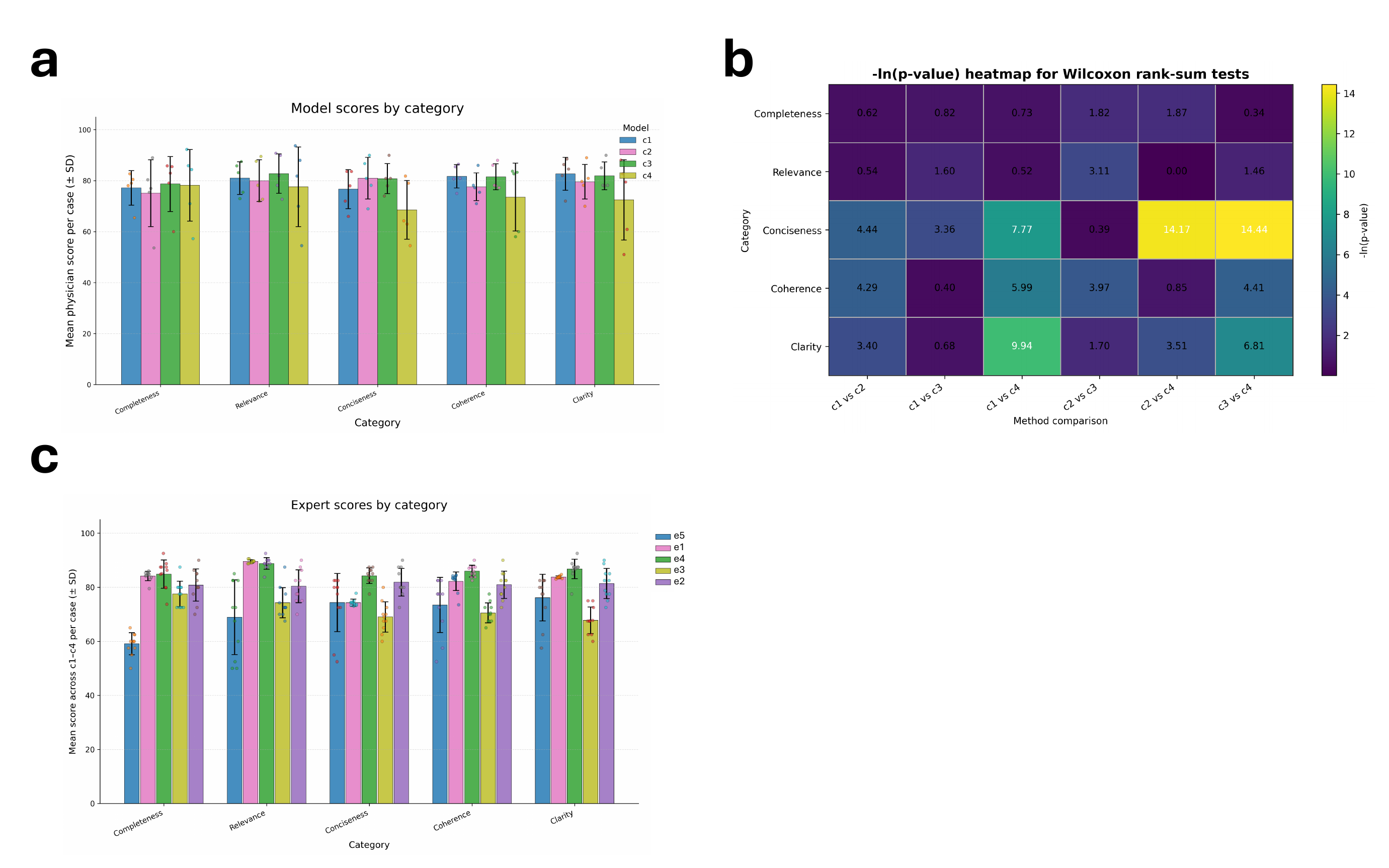}
    \caption{Additional information for pathology report generation experiments. (a) Evaluation scores by categories and models after aggregating all pathologists' results. (b) $-ln(p-value)$ based on the Wilcoxon Rank-Sum test across the scores of different methods. (c) Evaluation scores by categories and pathologists.}
    \label{supfig:addinforeport}
\end{figure}

\begin{figure}
    \centering
    \includegraphics[width=1\linewidth]{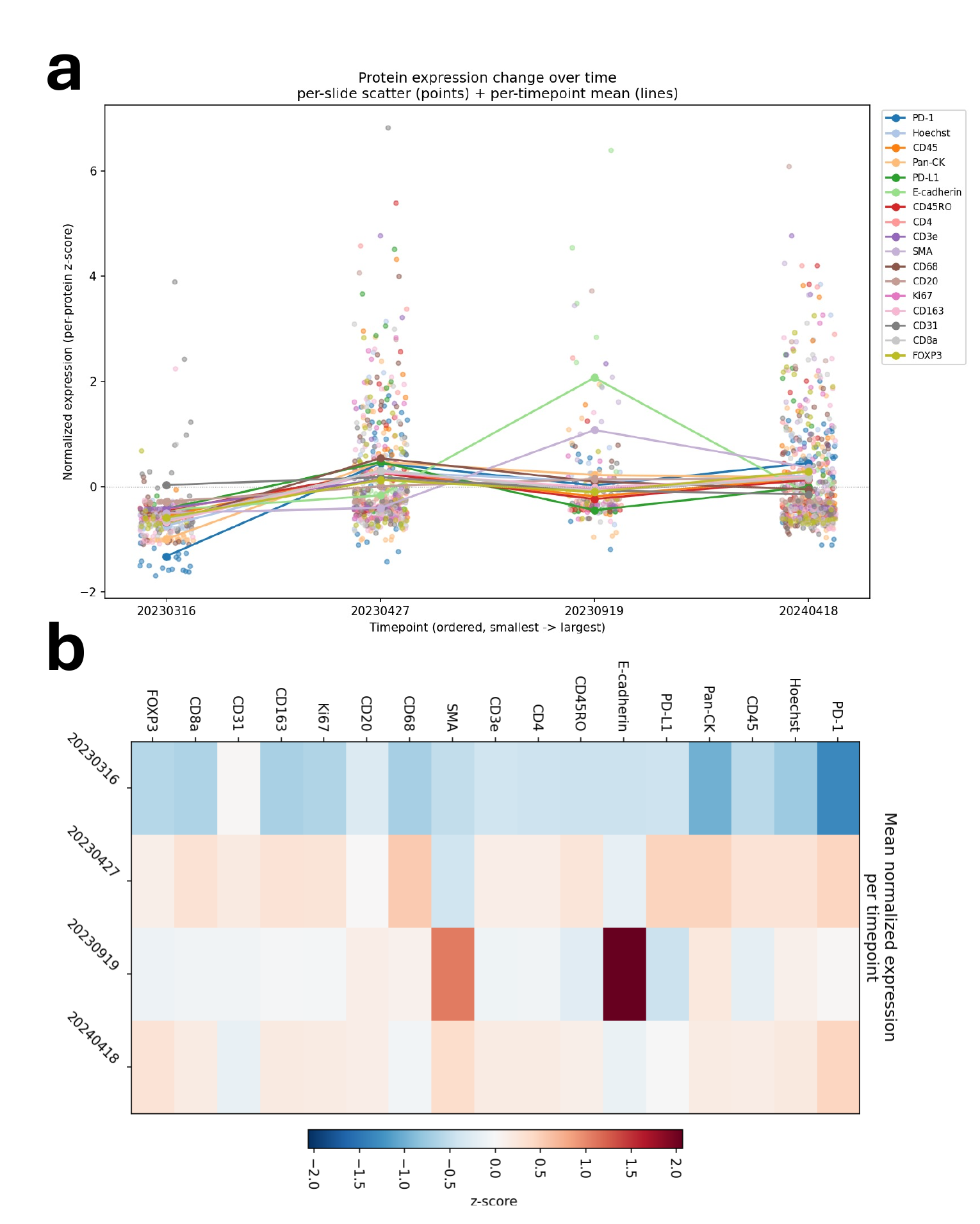}
    \caption{Case study for the joint modeling of mIF strength and EHR information across time points. (a) The changes of predicted mIF strength across time points and (b) Average expression levels across time points.}
    \label{supfig:another example}
\end{figure}

\end{document}